\pdfoutput=1
\PassOptionsToPackage{dvipsnames}{xcolor}
\documentclass[11pt]{article}

\usepackage[final]{acl}

\usepackage[utf8]{inputenc}
\usepackage{times}
\usepackage{dblfloatfix}
\usepackage{latexsym}
\usepackage{float}
\usepackage{todonotes}
\usepackage[most]{tcolorbox}
\usepackage{booktabs}
\usepackage{amsmath}
\usepackage{titlefoot}
\usepackage{pifont}
\usepackage{float}  
\usepackage{marvosym}
\usepackage{authblk}
\usepackage{silence}
\usepackage{amssymb}
\makeatletter
\@ifundefined{bibsep}{}{\setlength{\bibsep}{0pt}}
\makeatother

\usepackage[T1]{fontenc}
\usepackage[utf8]{inputenc}

\usepackage{siunitx}
\usepackage{microtype}
\usepackage{amsfonts}
\usepackage{cancel}
\usepackage{soul}
\usepackage{bm}
\usepackage{algorithm}
\usepackage{algorithmic}
\usepackage{xspace}
\usepackage{fontawesome}
\usepackage{mathrsfs}
\usepackage{multirow}
\usepackage{colortbl}
\usepackage{subfig}
\usepackage{graphicx}
\usepackage{tabularx}
\usepackage{arydshln}
\usepackage{makecell}
\usepackage{CJKutf8}
\usepackage{adjustbox}
\usepackage[dvipsnames]{xcolor}
\usepackage[utf8]{inputenc}
\usepackage{textcomp}

\newcolumntype{L}[1]{>{\raggedright\arraybackslash}p{#1}}
\newcolumntype{C}[1]{>{\centering\arraybackslash}m{#1}}
\newcolumntype{R}[1]{>{\raggedleft\arraybackslash}p{#1}}

\title{SurgiQ: A Large-Scale Multi-Domain Benchmark\\ for Evaluating Surgical Understanding in Large Language Models}

\author{Ayah Al-Naji, Edoardo Fazzari$^{*}$, Saif Alkindi,
Hamdan Alhadhrami, Preslav Nakov, Cesare Stefanini \\
Mohamed bin Zayed University of Artificial Intelligence, UAE \\
\{\texttt{ayah.al-naji, edoardo.fazzari, saif.alkindi,} \\
\texttt{hamdan.alhadhrami, preslav.nakov, cesare.stefanini}\} \texttt{@mbzuai.ac.ae}}

\begin{document}
\maketitle

\begin{abstract}
Reliable evaluation of large language models in surgery
remains underdeveloped. Broad medical benchmarks test
clinical knowledge, while surgery requires procedural
reasoning, management trade-offs, negation handling,
and selection among plausible operative decisions. We
present SurgiQ, a text-only, source-grounded benchmark of
13,055 four-option multiple-choice questions spanning six
surgical domains and four question formats: case-based,
reasoning, best-option, and negative. SurgiQ is constructed
from surgical textbooks, open-access papers, and examination
material using a multi-stage generation, verification, and
expert-audit pipeline. We evaluate 35 open-weight
LLMs under a unified log-likelihood protocol. Our results show
substantial remaining headroom: smaller models often remain
near the 25\% random baseline, while the best model reaches
68.1\% accuracy. General-purpose models, especially Qwen2.5,
outperform most biomedical models, suggesting that current
medical specialization does not yet provide sufficiently broad
surgical coverage. Calibration and error analysis further show
that even strong models make confident mistakes on clinically
plausible distractors, motivating more reliable and broader
surgical LLM evaluation.
\href{https://anonymous.4open.science/r/SurgiQ-6BDC/}{Our data and code are available at  \faGithub.}
\end{abstract}

\section{Introduction}

\begin{figure}[t]
    \centering
    \includegraphics[width=0.9\linewidth]{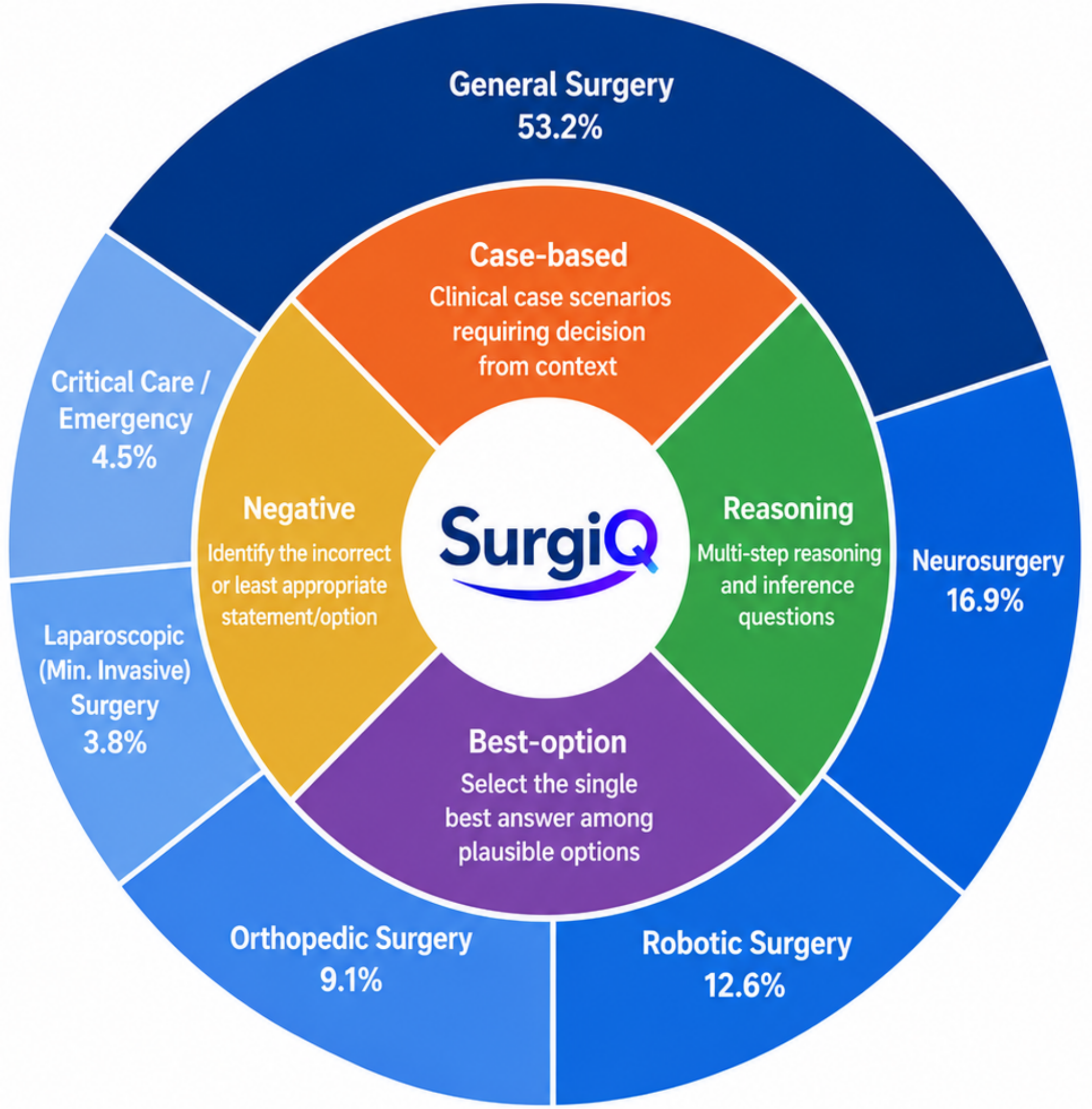}
    \caption{Overview of the SurgiQ benchmark, illustrating the diversity of surgical domains and question categories. The outer ring shows the distribution of questions across six surgical domains, while the inner ring illustrates the four MCQ formats used throughout the dataset. Each surgical domain contains a mixture of all four question types.}
    \label{fig:surgiq_overview}
\end{figure}
 
Recent advances in foundation models and reasoning-oriented
Large Language Models (LLMs) have accelerated interest in healthcare applications,
where reliable and well-calibrated decision-making is
critical. Rigorous evaluation has therefore become essential,
motivating challenging knowledge-intensive benchmarks across
general reasoning and medicine~\cite{hendrycks2020measuring, rein2023gpqa, wang2024mmlu}.

Medical benchmarks such as MedQA-USMLE~\cite{jin2021disease},
MedMCQA~\cite{pal2022medmcqa}, PubMedQA~\cite{jin2019pubmedqa},
and MultiMedQA~\cite{singhal2023large} evaluate broad clinical
and biomedical knowledge. Specialty-oriented resources such as
MedExQA~\citep{kim-etal-2024-medexqa} further show that
medical coverage is uneven across subfields. However, surgery
poses additional challenges~\cite{fazzari2026survey}: procedural sequencing, operative
anatomy, intraoperative judgment, perioperative management,
and choosing the best option among clinically plausible
alternatives.

Surgical AI evaluation has recently advanced through multimodal
benchmarks for surgical scene understanding, including
SurgMLLMBench~\citep{choi2025surgmllmbench} and
SurgVLM-Bench~\citep{zeng2025surgvlm}. While these benchmarks
provide valuable evaluation of visual perception, workflow,
instrument use, and video/image understanding, they do not target
source-grounded textual assessment of surgical knowledge and
operative decision-making across specialties. SurgiQ fills this
complementary gap by evaluating text-based surgical question
answering and calibration in open-weight LLMs.

Here, we introduce \textbf{SurgiQ}, a large-scale multiple-choice
benchmark comprising 13,055 clinically grounded questions
across six surgical domains and four question formats. We
evaluate 35 open-weight general-purpose, reasoning-oriented,
and biomedical LLMs under a unified zero-shot and few-shot
framework. Our focus on open-weight models supports
reproducible local evaluation and is practically relevant for
clinical or robotic systems where connectivity, privacy, and
latency constraints may limit reliance on external APIs. SurgiQ
is not a deployment test; rather, it measures a prerequisite
capability: controlled text-based discrimination among surgical
decisions. Figure~\ref{fig:surgiq_overview} provides an
overview of the benchmark structure.
 
We make the following contributions:
 
\begin{itemize}
 
    \item \textbf{Dataset.} SurgiQ provides 13,055 source-grounded surgical MCQs across six domains and four question types (case-based, reasoning, best-option, and negative), with per-question metadata and a versioned public release.
 
    \item \textbf{Validation.} We establish benchmark quality through a physician audit, a human surgeon baseline, near-duplicate analysis, answer-order robustness tests, and shortcut baselines, collectively confirming that SurgiQ cannot be solved by surface-level heuristics.
 
    \item \textbf{Evaluation.} We assess 35 open-weight LLMs spanning general-purpose, biomedical, and reasoning-oriented families under a unified, deterministic log-likelihood scoring protocol, with complementary calibration, domain, question-type, and error analyses.
 
    \item \textbf{Findings.} General-purpose models outperform most medical-specialized models on surgical reasoning; calibration analysis and distractor-level error breakdowns reveal that confident failures on clinically plausible alternatives remain a major reliability gap across model families.
 
\end{itemize}

\section{Related Work}

General benchmarks such as MMLU~\cite{hendrycks2020measuring},
BIG-bench Hard~\cite{suzgun2023challenging}, GPQA~\cite{rein2023gpqa},
and MMLU-Pro~\cite{wang2024mmlu} established large-scale evaluation
for academic and expert reasoning. Medical benchmarks then extended
this paradigm to clinical settings: MedQA-USMLE~\cite{jin2021disease},
PubMedQA~\cite{jin2019pubmedqa}, MedMCQA~\cite{pal2022medmcqa}, and
MultiMedQA~\cite{singhal2023large} evaluate biomedical knowledge and
clinical reasoning, while specialty resources such as
MedExQA~\citep{kim-etal-2024-medexqa} highlight uneven coverage across
medical subfields and the need for richer specialty evaluation.

However, surgery remains only partially covered by existing medical
QA resources. Surgical reasoning requires operative anatomy,
procedural sequencing, perioperative management, negation handling,
and context-dependent selection among plausible interventions.
Recent surgical benchmarks have primarily expanded evaluation along
the multimodal axis. For example, SurgMLLMBench~\citep{choi2025surgmllmbench}
integrates surgical VQA, workflow annotations, and instrument
segmentation across laparoscopic, robot-assisted, and microsurgical
settings, while SurgVLM-Bench~\citep{zeng2025surgvlm} evaluates
vision-language models on surgical perception, temporal understanding,
and scene-level reasoning. These benchmarks are valuable for
intraoperative visual understanding, but they do not directly evaluate
source-grounded textual surgical knowledge, examination-style reasoning,
or calibrated answer selection across surgical specialties. SurgiQ
therefore targets a complementary evaluation axis: broad text-based
surgical QA and clinically grounded operative decision-making using
open-weight LLMs.

Table~\ref{tab:benchmark_comparison} summarizes the positioning of SurgiQ relative to representative general, medical, and surgical benchmarks. Existing medical QA benchmarks primarily evaluate broad biomedical knowledge,
while recent surgical benchmarks focus mainly on multimodal perception and
workflow understanding. In contrast, SurgiQ uniquely combines specialized surgical knowledge evaluation, text-only evaluation, calibration analysis, and large-scale evaluation of open-weight LLMs within a unified benchmark.

\begin{table*}[t]
\centering
\small
\setlength{\tabcolsep}{5pt}
\begin{tabular}{llcccc}
\toprule
Benchmark & \# Questions & Format & Med. QA & Surg. Focus & Calibration \\
\midrule
MMLU~\cite{hendrycks2020measuring}        & 15,908 & MCQ & $\times$ & $\times$ & $\times$ \\
MedQA-USMLE~\cite{jin2021disease}         & 12,723 & MCQ & \checkmark & $\times$ & $\times$ \\
PubMedQA~\cite{jin2019pubmedqa}           & 1,000 labeled & Yes/no/maybe & \checkmark & $\times$ & $\times$ \\
MedMCQA~\cite{pal2022medmcqa}             & 194k+ & MCQ & \checkmark & $\times$ & $\times$ \\
MedExQA~\citep{kim-etal-2024-medexqa}     & 965 & MCQ  & \checkmark & $\times$ & $\times$ \\
\textbf{SurgiQ (ours)}                    & \textbf{13,055} & MCQ & \checkmark & \checkmark & \checkmark \\
\bottomrule
\end{tabular}
\caption{Positioning of SurgiQ relative to representative text-based QA datasets.}
\label{tab:benchmark_comparison}
\end{table*}

\begin{table}[t]
\centering
\small
\begin{tabular}{lrr}
\toprule
\textbf{Surgical Domain} & \textbf{\# MCQs} & \textbf{\%} \\
\midrule
General Surgery           & 6,948 & 53.2\% \\
Neurosurgery              & 2,204 & 16.9\% \\
Robotic Surgery           & 1,643 & 12.6\% \\
Orthopedic Surgery        & 1,189 &  9.1\% \\
Critical Care / Emergency &   590 &  4.5\% \\
Laparoscopic Surgery      &   493 &  3.8\% \\
\midrule
\textbf{Total} & \textbf{13,055} & \textbf{100\%} \\
\bottomrule
\end{tabular}
\caption{SurgiQ domain distribution across six surgical specialties.}
\label{tab:domain-distribution}
\end{table}

\section{SurgiQ}

\begin{figure}[t]
\centering
\includegraphics[width=\columnwidth]{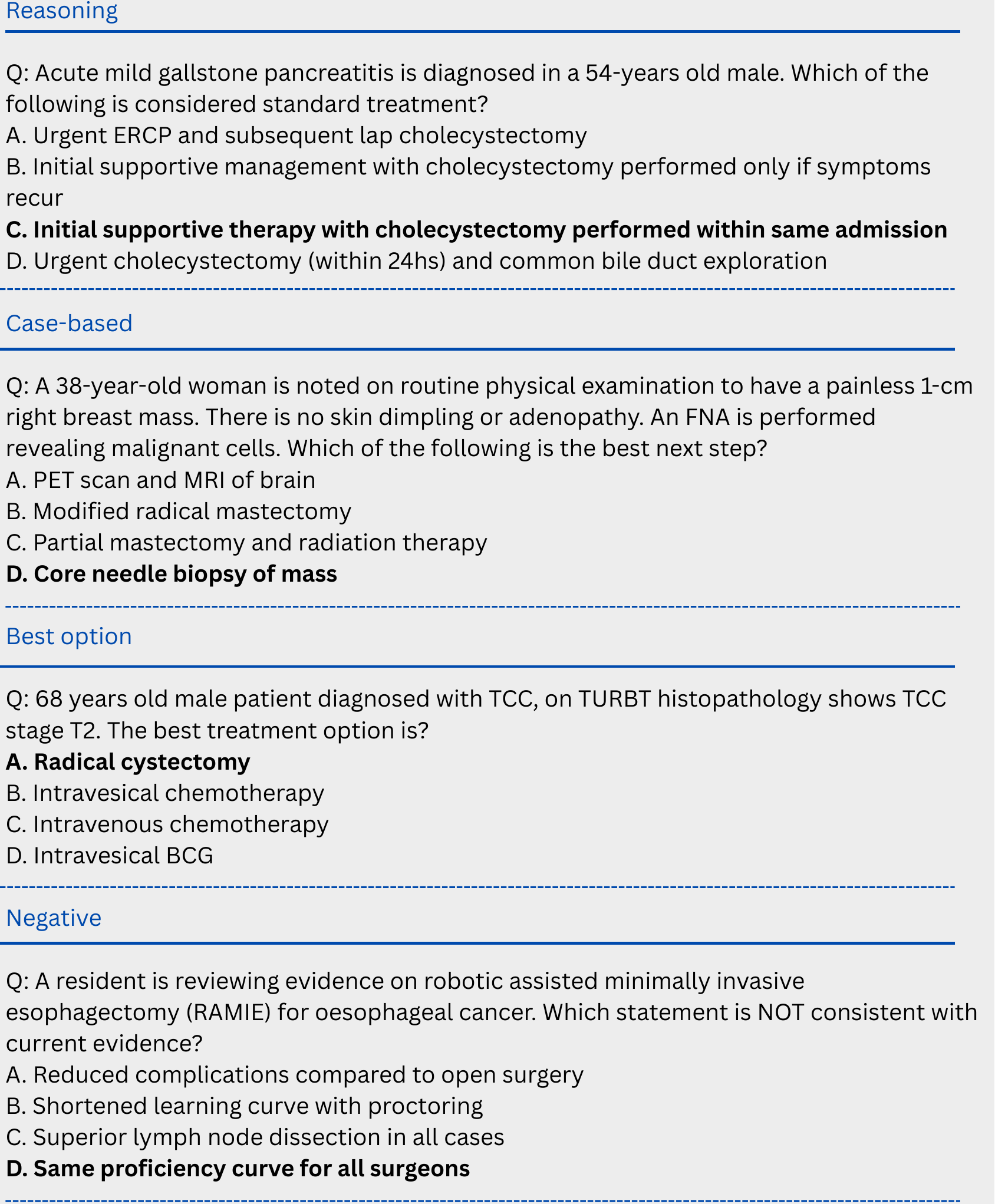}
\caption{Examples of SurgiQ questions across four formats: reasoning, case-based, best-option, and negative. The correct answer in each example is shown in bold.}
\label{fig:surgiq_examples}
\end{figure}

\begin{figure*}[t]
\centering
\includegraphics[width=0.95\textwidth]{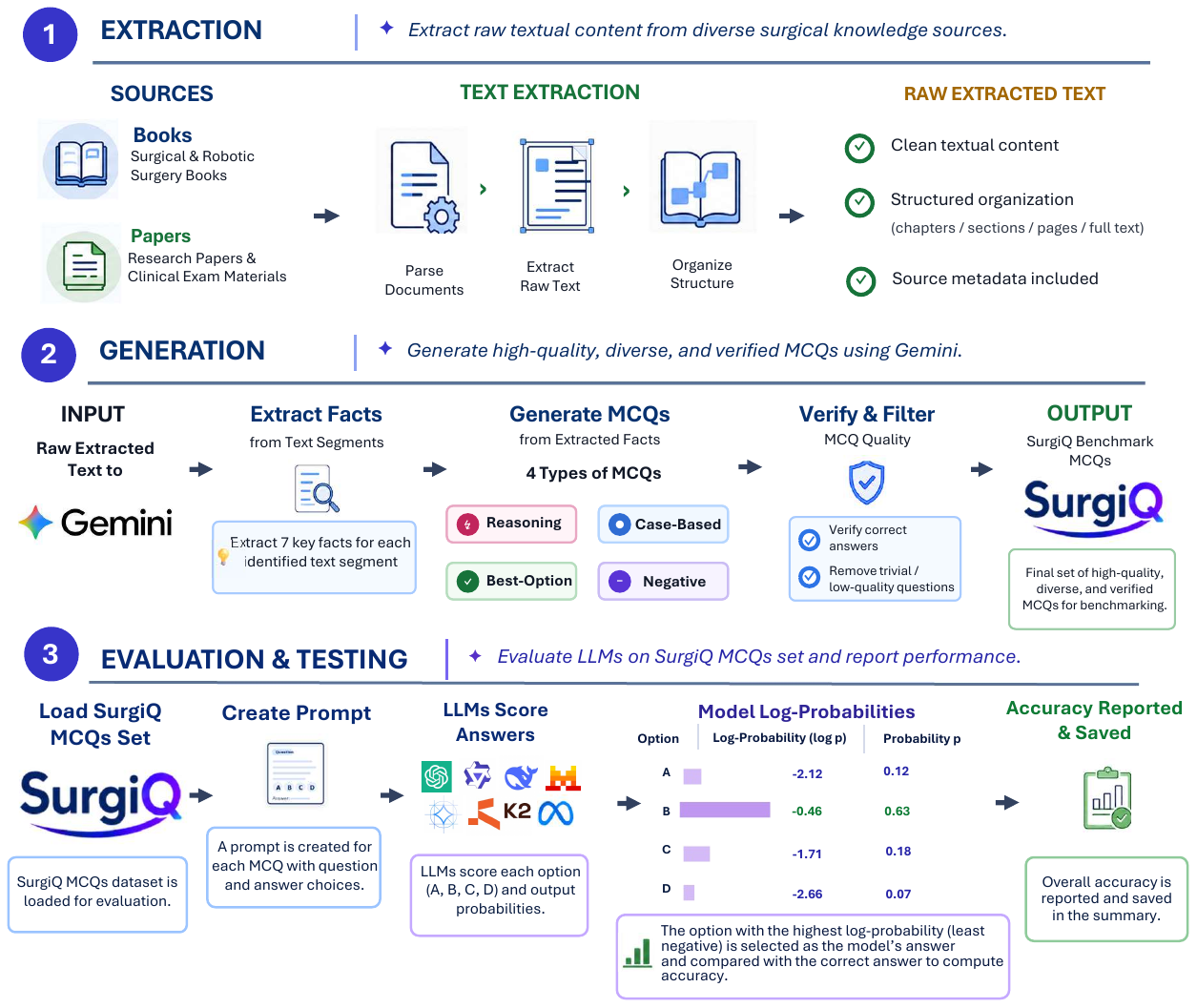}
\caption{Overview of the SurgiQ construction pipeline.}
\label{fig:pipeline}
\end{figure*}

We introduce SurgiQ, a surgical multiple-choice question-answering dataset for evaluating reasoning and decision-making in surgical contexts. SurgiQ contains 13,055 questions spanning six surgical domains: general surgery, neurosurgery, robotic surgery, orthopedic surgery, critical care, and laparoscopic surgery. We construct the dataset from 27 surgical textbooks, 635 open-access research papers, and a small collection of surgical examination materials. Appendix~\ref{sec:surgical-reference-books} lists the textbook sources, while the open-access papers are omitted from the appendix due to space constraints. Each question includes four candidate answers, one correct option, and associated metadata. Table~\ref{tab:domain-distribution} summarizes the domain coverage across surgical specialties, while Figure~\ref{fig:surgiq_examples} presents representative examples of the four question formats included in SurgiQ.

\subsection{Data Construction}

We organize books into chapter-level JSON files and split them into page-level segments, while we process research papers using their full text. We associate each source with metadata to support consistent downstream processing across formats. During preprocessing, we retain only textual content and remove non-textual elements such as images and tables. We then process the source documents using format-specific extraction pipelines to recover structured textual content suitable for downstream question generation.

\subsection{Question Generation}
\label{sec:question_generation}

We generate SurgiQ questions using a prompt-based pipeline built on top of Gemini 2.5 Flash~\citep{comanici2025gemini25}. As illustrated in Figure~\ref{fig:pipeline}, the pipeline first extracts high-yield facts from source text segments and then generates clinically grounded multiple-choice questions. We design the prompts to encourage multi-step reasoning, avoid trivial recall, and produce plausible distractors. 

We use separate fact-extraction prompts for books and research papers, while question generation follows a unified template across all source types. Appendix~\ref{sec:prompt_design} provides the prompt templates used for fact extraction, question generation, verification, filtering, and evaluation. Each question includes four candidate options (A--D), one correct answer, and a concise explanation. SurgiQ contains four question formats: reasoning, case-based, best-option, and negative questions.

We link each generated question to its source text segment to ensure traceability. We perform question generation, verification, and quality filtering within a unified Gemini 2.5 Flash pipeline using task-specific temperatures (0.1 for fact extraction and verification; 0.35 for question generation). Pipeline accounting indicates that the initial generation stage produced 17,068 candidate MCQs from 7,216 extracted textbook segments and 659 research papers. After answer verification, Gemini-based quality filtering, rule-based checks, deduplication, and manual cleanup, 4,013 candidates (23.5\%) were removed, yielding the final 13,055-question release. Rejection categories included unsupported source grounding, ambiguous or multiple plausible answers, malformed options, trivial recall, purely lexical cues, duplicate or near-duplicate items, and answer-label inconsistencies. 

We shuffle answer choices to balance correct-answer positions (A: 26.9\%, B: 23.8\%, C: 24.7\%, D: 24.6\%). Appendix Figure~\ref{fig:evaluation_prompt_example} presents the evaluation prompt used during inference-time benchmarking and model evaluation in SurgiQ.

Using Gemini 2.5 Flash for both generation and verification introduces a potential circularity concern. We mitigate this risk through rule-based filtering, independent expert validation, and evaluation on exclusively non-Gemini open-weight models. We manually design all prompts and iteratively refine them through pilot generation experiments.

\subsection{Quality Control}
To assess dataset quality, 300 randomly sampled
SurgiQ questions were independently reviewed by
three physicians for answer correctness and
question clarity. Annotators evaluated whether
the designated correct answer was medically
accurate and whether the question was clear and
clinically interpretable. Using majority
consensus labels, 92\% of audited items
contained correct answers and 90\% were judged
clear and unambiguous. Inter-annotator
agreement was substantial to near-perfect, with
Fleiss' $\kappa = 0.84$ for answer correctness
and $\kappa = 0.72$ for clarity assessment.
Most disagreements arose in questions involving
institution-dependent perioperative practices
or nuanced operative decision-making scenarios.

\subsection{Dataset Statistics}

The final SurgiQ dataset contains 13,055 multiple-choice questions spanning six surgical domains and four question formats. As shown in Table~\ref{tab:question-types}, case-based questions form the largest portion of the dataset (47.1\%), followed by reasoning (20.5\%), negative (16.5\%), and best-option questions (16.0\%). This distribution emphasizes clinically grounded decision-making and procedural reasoning. Table~\ref{tab:domain-distribution} summarizes the distribution across surgical domains, with General Surgery representing the largest portion of the dataset (53.2\%). We include Neurosurgery because of its overlap with broader surgical procedural reasoning and clinical decision-making. Most questions originate from surgical textbooks (87.6\%), while the remainder come from research papers (10.0\%) and clinical examination materials (2.3\%). The domain distribution shown in Table~\ref{tab:domain-distribution} partially reflects the natural availability of accessible surgical literature across specialties rather than deliberate balancing during data collection. The average question length is 59.8 words (median: 61; std: 26.6), indicating moderate variation in question complexity. Appendix~\ref{tab:detailed_domain_type_stats} provides a more detailed breakdown across domains and question types.
\begin{table}[t]
\centering
\small
\begin{tabular}{lrr}
\toprule
\textbf{Question Type} & \textbf{\# MCQs} & \textbf{\%} \\
\midrule
Case-based   & 6,156 & 47.1\% \\
Reasoning    & 2,673 & 20.5\% \\
Negative     & 2,153 & 16.5\% \\
Best-option  & 2,085 & 16.0\% \\
\midrule
\textbf{Total} & \textbf{13,055} & \textbf{100\%} \\
\bottomrule
\end{tabular}
\caption{Questions distribution by question type.}
\label{tab:question-types}
\end{table}

\section{Experiments}

\subsection{Setup}

We evaluate SurgiQ in a zero-shot multiple-choice setting using a diverse set of open-weight large language models, including general-purpose, biomedical, and reasoning-oriented architectures. We evaluate models spanning a wide range of parameter scales and model families, including Qwen2.5~\citep{qwen2024qwen25}, LLaMA-based models~\citep{touvron2023llama}, Mistral-family models~\citep{jiang2023mistral}, and DeepSeek-R1-Distill~\citep{deepseek2025r1}. Appendix Table~\ref{tab:model_artifacts} provides the full list of evaluated models and their corresponding repositories. We focus on open-weight models to enable reproducible local evaluation and to avoid dependence on external API availability, latency, and versioning changes during benchmarking. This choice is also relevant to clinical and robotic settings where local inference may be preferable for privacy, reliability, and connectivity reasons~\cite{amparore2024computer, fazzari2026real}. Following standard practice in LLM benchmark evaluation~\citep{hendrycks2020measuring,rein2023gpqa,wang2024mmlu}, we evaluate the models using zero-shot prompting, as shown in Figure~\ref{fig:evaluation_prompt}, and use accuracy as the primary metric. In addition, we perform an exploratory few-shot analysis following prior work on in-context learning~\citep{NEURIPS2020_1457c0d6} to examine how limited task demonstrations influence model performance and prompt sensitivity.

\begin{figure}[t]
\centering
\fbox{
\begin{minipage}{0.92\columnwidth}
\small
\ttfamily
Select the correct answer.\\[0.6em]

Question: [QUESTION]\\
A. [OPTION A]\\
B. [OPTION B]\\
C. [OPTION C]\\
D. [OPTION D]\\[0.6em]

Answer:
\end{minipage}
}
\caption{Zero-shot prompt template used for model evaluation. The placeholders \textsc{[QUESTION]} and \textsc{[OPTION]} are replaced with the corresponding question and answer choices.}
\label{fig:evaluation_prompt}
\end{figure}
\begin{table*}[t]
\centering
\small
\setlength{\tabcolsep}{0pt}
\renewcommand{\arraystretch}{0.95}

\begin{tabular*}{\textwidth}{@{\extracolsep{\fill}}lccccc@{}}
\toprule
\textbf{Model} & \textbf{Reasoning} & \textbf{Case-based} & \textbf{Best-option} & \textbf{Negative} & \textbf{Avg.} \\
\midrule

Random & 25.00 & 25.00 & 25.00 & 25.00 & 25.00 \\
\hdashline

Gemma-2-2B-IT~\citep{gemmateam2024gemma2} & 36.26 & 35.79 & 32.84 & 33.09 & 34.97 \\
Gemma-2-9B-IT & 44.41 & 43.60 & 41.81 & 46.10 & 43.87 \\
MedGemma-4B-IT~\citep{medgemma2025} & 48.61 & 45.26 & 47.00 & 32.39 & 44.08 \\
MedGemma-27B-IT & 40.35 & 39.45 & 36.25 & 36.20 & 38.58 \\
\hdashline

BioMistral-7B~\citep{labrak2024biomistral} & 50.53 & 47.53 & 47.05 & 41.87 & 47.14 \\
\hdashline

Llama3-OpenBioLLM-8B~\citep{dubey2024llama3} & 53.04 & 52.36 & 50.31 & 48.98 & 51.60 \\
Llama3-Med42-8B~\citep{dubey2024llama3} & 57.47 & 54.49 & 53.67 & 56.27 & 55.24 \\
\hdashline

TxGemma-2B~\citep{txgemma2025} & 25.19 & 24.67 & 25.68 & 25.28 & 25.02 \\
TxGemma-9B-Predict & 26.84 & 25.13 & 24.53 & 28.86 & 25.99 \\
TxGemma-9B-Chat & 29.09 & 28.24 & 27.27 & 31.78 & 28.82 \\
TxGemma-27B-Predict & 32.81 & 32.63 & 30.92 & 33.41 & 32.51 \\
TxGemma-27B-Chat & 56.94 & 54.72 & 56.07 & 54.88 & 55.40 \\
\hdashline

K2-Think-V2~\citep{k2team2025k2v2} & 60.21 & 58.20 & 55.11 & 46.28 & 56.15 \\
\hdashline

Meditron-7B~\citep{chen2023meditron} & 27.06 & 27.54 & 27.03 & 26.16 & 27.14 \\
Meditron-70B & 61.22 & 58.12 & 57.42 & 45.17 & 56.50 \\
\hdashline

DeepSeek-R1-Distill-Qwen-7B~\citep{deepseek2025r1} & 34.50 & 33.90 & 33.70 & 34.15 & 34.02 \\
DeepSeek-R1-Distill-Qwen-14B & 57.81 & 55.27 & 51.90 & 59.25 & 55.92 \\
DeepSeek-R1-Distill-Qwen-32B & 61.56 & 59.90 & 55.98 & 64.36 & 60.37 \\
\hdashline

Ministral-3-8B-Instruct~\citep{ministral2026} & 60.17 & 55.75 & 54.58 & 43.12 & 54.40 \\
Ministral-3-8B-Reasoning & 60.62 & 57.96 & 56.02 & 59.48 & 58.45 \\
Ministral-3-14B-Reasoning & 65.69 & 63.22 & 59.91 & 64.96 & 63.48 \\
\hdashline

MedMO-4B~\citep{deria2026medmo} & 63.66 & 61.38 & 59.34 & 60.22 & 61.33 \\
MedMO-4B-Next & 63.69 & 61.41 & 59.31 & 60.24 & 61.35 \\
MedMO-8B-Next & 65.65 & 62.42 & 60.92 & 60.22 & 62.47 \\
MedMO-8B & 67.23 & 63.54 & 60.30 & 62.73 & 63.64 \\
\hdashline

GPT-OSS-20B~\citep{openai2025gptoss} & 62.16 & 58.85 & 56.65 & 48.98 & 57.56 \\
GPT-OSS-120B & 67.91 & 64.75 & 62.70 & 62.08 & 64.63 \\
\hdashline

Meerkat-8B~\citep{kim2025meerkat} & 57.92 & 56.05 & 54.10 & 55.76 & 56.09 \\
Meerkat-70B & 67.12 & 64.97 & 61.35 & 67.15 & 65.19 \\
\hdashline

HuatuoGPT-o1-7B~\citep{chen2024huatuogpt} & 58.86 & 58.27 & 55.88 & 57.67 & 57.92 \\
HuatuoGPT-o1-70B & 68.21 & 65.71 & 60.87 & 65.75 & 65.45 \\
\hdashline

Qwen2.5-32B-Instruct~\citep{qwen2024qwen25} & 67.15 & 64.23 & 61.88 & 67.01 & 64.92 \\
Qwen2.5-32B & 67.45 & 65.98 & 61.74 & 64.82 & 65.43 \\
Qwen2.5-72B-Instruct & \textbf{71.25} & 67.38 & \textbf{65.58} & 68.03 & 68.00 \\
Qwen2.5-72B & 70.83 & \textbf{67.90} & 64.76 & \textbf{68.31} & \textbf{68.08} \\

\bottomrule
\end{tabular*}

\caption{Zero-shot LLM performance (\%) on SurgiQ across question types after duplicate filtering. Model families are grouped together and ordered in ascending order according to the highest-performing model within each family. Random denotes uniform guessing; Avg. is micro-accuracy over all SurgiQ items.}
\label{tab:results_by_type}
\vspace{-1em}
\end{table*}

\subsection{Implementation Details}

We evaluate all models under a unified zero-shot multiple-choice setting. For each question, models receive a fixed prompt template and must select one of the candidate answers. For open-weight models, predictions are computed from next-token log-probabilities over the answer options. To reduce tokenization artifacts, we evaluate both plain and space-prefixed variants of each option label, use the maximum log-probability across variants as the final score~\citep{geh-etal-2024-signal}, and derive confidence estimates by applying a softmax over the candidate logits for calibration analysis. To mitigate option-position bias, answer choices are randomly shuffled once during dataset construction and the resulting order is kept fixed across all evaluations. Because predictions are obtained directly from log-probabilities rather than autoregressive sampling, the evaluation procedure is fully deterministic across runs.

\subsection{Results}

\textbf{Results across all models.}
Table~\ref{tab:results_by_type} reports accuracy across all evaluated models and question types. Qwen2.5-72B~\citep{qwen2024qwen25} achieves the highest overall accuracy, followed by HuatuoGPT-o1-70B~\citep{chen2024huatuogpt} and Meerkat-70B~\citep{kim2025meerkat}. Appendix~\ref{tab:all_models_domain_accuracy} provides detailed domain-level accuracy results across all surgical specialties. Across model families, larger models generally outperform smaller ones. For example, DeepSeek-R1-Distill-Qwen-32B~\citep{deepseek2025r1} substantially outperforms its 7B counterpart, with similar trends appearing for Qwen2.5 and Meerkat. This pattern suggests that capacity and base-model strength play important roles in surgical reasoning. At the same time, medical specialization alone remains insufficient. Although medically-oriented models such as HuatuoGPT-o1~\citep{chen2024huatuogpt} and medical multimodal models such as MedMO~\citep{deria2026medmo} remain competitive, general-purpose models such as Qwen2.5~\citep{qwen2024qwen25} achieve the strongest overall performance. This finding suggests that broad reasoning ability and knowledge coverage may be as important as biomedical adaptation for surgical MCQ performance. Smaller models (e.g., 2B--9B parameters) often perform near random chance, highlighting both the difficulty of SurgiQ and the limitations of smaller-scale architectures for reliable clinical reasoning. Overall, the substantial performance gap across models indicates considerable room for improvement. 
Additional evaluation on a manually audited 300-question subset shows that surgeons achieved 89.1\% accuracy, substantially outperforming the best evaluated open-weight LLM on the same subset (51.3\%), indicating that SurgiQ remains clinically nontrivial and far from saturated; full results are reported in Appendix Table~\ref{tab:audited_300_baselines}.

\textbf{Results across question types.}
Performance varies substantially across question types, revealing model-dependent reasoning patterns rather than a uniform difficulty ordering. Best-option questions consistently rank among the most challenging because multiple answer choices are often partially correct while only one remains optimal within the specific clinical context. Negative questions additionally exhibit high variance across models, suggesting substantial differences in how architectures handle logical negation. Moreover, reasoning questions are not uniformly harder than case-based questions. Stronger models such as Qwen2.5-72B~\citep{qwen2024qwen25} perform better on reasoning questions than on case-based questions, whereas smaller models remain near chance across all formats. This pattern suggests that the ability to differentiate question types emerges primarily at larger model scales. Overall, these findings demonstrate that question difficulty depends strongly on both reasoning format and model capability, highlighting the value of SurgiQ for fine-grained evaluation of clinical reasoning behavior.

\subsection{Analysis}

\begin{figure}[t]
\centering
\includegraphics[width=0.85\columnwidth]{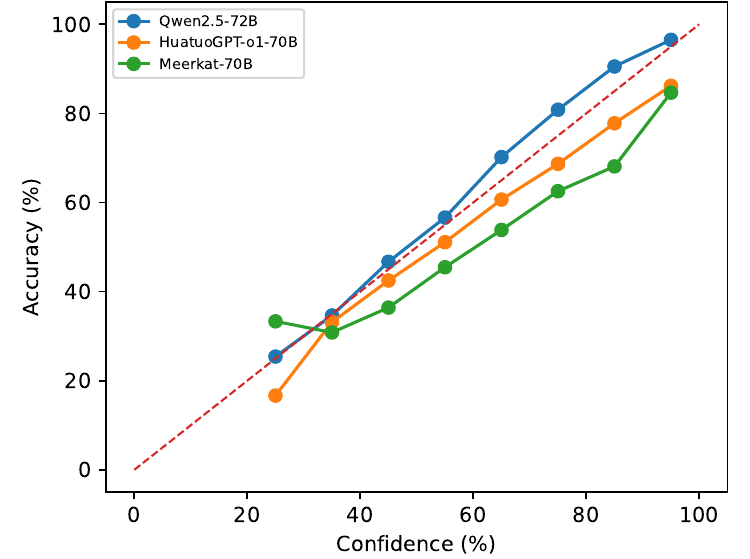}
\caption{Reliability diagram showing model calibration on SurgiQ. The diagonal line represents perfect calibration.}
\label{fig:calibration}
\end{figure}

\textbf{Model calibration.}
We analyze calibration for the three strongest model families by comparing prediction confidence with empirical accuracy (Figure~\ref{fig:calibration}). Table~\ref{tab:calibration} reports top-label Expected Calibration Error (ECE), Brier score, and confidence--accuracy gap computed from the four-option probabilities. Qwen2.5-72B~\citep{qwen2024qwen25} shows the smallest mismatch and remains mildly underconfident (gap = $-$0.033), whereas Meerkat-70B~\citep{kim2025meerkat} is substantially overconfident (ECE = 0.120). The observed accuracy--calibration relationship likely arises because SurgiQ errors are highly discriminative: stronger models better separate the correct option from plausible distractors, while less aligned models may assign high confidence to familiar but incorrect clinical choices. The remaining calibration gaps further suggest that model confidence is not a reliable proxy for surgical certainty.

\begin{table}[t]
\centering
\small
\begin{tabular}{lrrr}
\toprule
\textbf{Model} & \textbf{ECE} & \textbf{Brier} & \textbf{Gap} \\
\midrule
Qwen2.5-72B      & 0.036 & 0.181 & $-$0.033 \\
HuatuoGPT-o1-70B & 0.058 & 0.204 & $+$0.058 \\
Meerkat-70B      & 0.120 & 0.214 & $+$0.120 \\
\bottomrule
\end{tabular}
\caption{Calibration metrics for representative top-performing
models. ECE = Expected Calibration Error; Gap = mean
confidence minus mean accuracy.}
\label{tab:calibration}
\end{table}

\begin{figure}[t]
\centering
\includegraphics[width=\columnwidth]{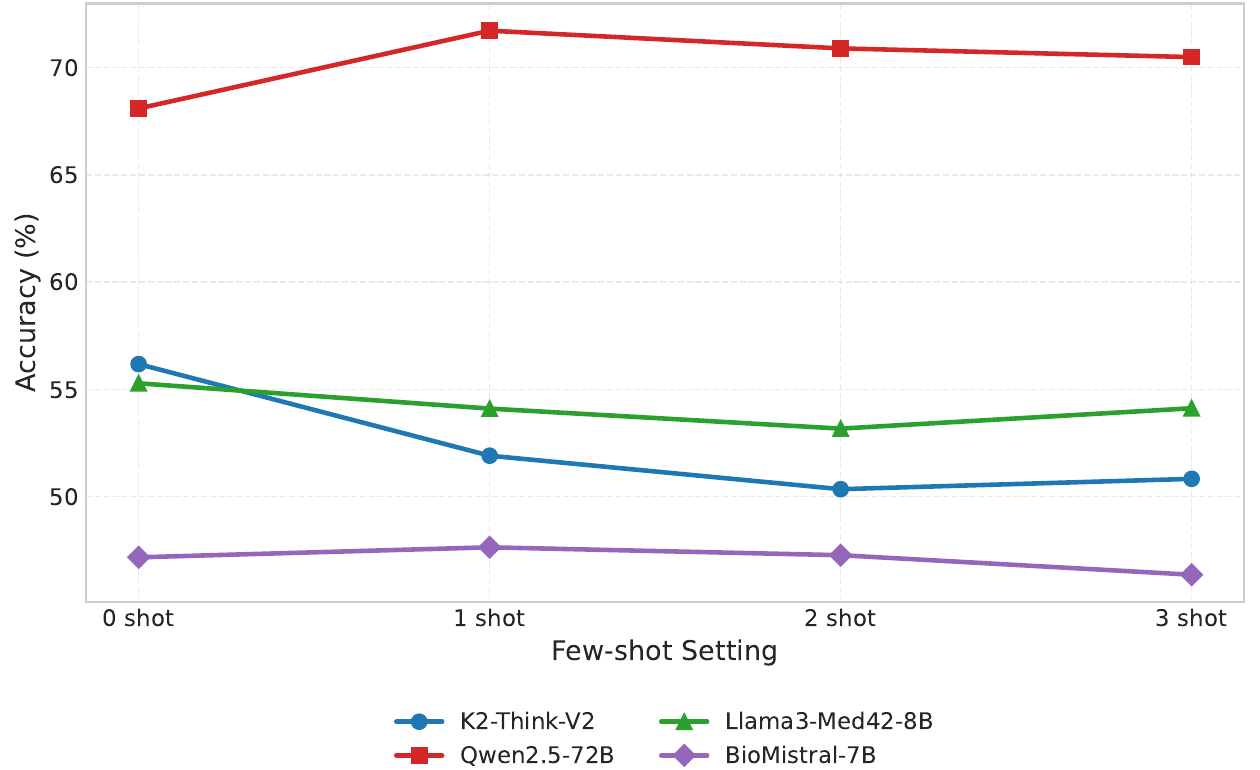}
\caption{Zero-shot and few-shot performance on SurgiQ.}
\label{fig:fewshot}
\end{figure}

\textbf{Few-shot prompting.}
We conduct an exploratory in-context learning study using four representative models under 1-, 2-, and 3-shot settings (Figure~\ref{fig:fewshot}). Qwen2.5-72B~\citep{qwen2024qwen25} improves from 68.1\% (0-shot) to 72.1\% with one example, although additional examples do not provide consistent gains. This improvement likely reflects format alignment rather than new surgical knowledge: a single demonstration clarifies the A--D selection protocol and enables Qwen2.5-72B's broad pretrained knowledge to transfer more effectively to the task. Additional examples may instead introduce lexical priors or prompt noise. In contrast, K2-Think-V2~\citep{k2team2025k2v2} degrades under few-shot prompting, while Llama3-Med42-8B~\citep{dubey2024llama3} and BioMistral-7B~\citep{labrak2024biomistral} remain close to their zero-shot performance. Overall, these results suggest that the few-shot setting primarily reflects prompt sensitivity rather than a fully optimized in-context learning protocol.

\textbf{Domain difficulty.}
Table~\ref{tab:domain-means} reports the mean accuracy per domain across all 35 evaluated models. Orthopedic surgery is the lowest-scoring domain (49.23\% $\pm$ 11.24\%) and performs significantly below every other domain under a Wilcoxon signed-rank test ($p < 0.001$), while laparoscopic surgery scores below neurosurgery ($p = 0.008$). We interpret these tests descriptively because the domain and question-type distributions are not balanced. Orthopedics contains a larger proportion of reasoning and best-option questions and relies on narrower source material (Appendix Table~\ref{tab:detailed_domain_type_stats}), suggesting that the observed gap may reflect both specialty-specific content and item-format effects.
\begin{table}[t]
\centering
\small
\setlength{\tabcolsep}{6pt}
\begin{tabular}{lrr}
\toprule
\textbf{Domain} & \textbf{Mean (\%)} & \textbf{Std.} \\
\midrule
Neurosurgery              & 53.08 & 13.20 \\
Robotic Surgery           & 52.67 & 12.97 \\
General Surgery           & 52.57 & 13.77 \\
Critical Care / Emergency & 52.45 & 12.63 \\
Laparoscopic Surgery      & 52.03 & 12.49 \\
Orthopedic Surgery        & \textbf{49.23} & \textbf{11.24} \\

\bottomrule
\end{tabular}
\caption{Mean accuracy per domain across 35 models.}
\label{tab:domain-means}
\end{table}

\textbf{Failure patterns.}
As a qualitative case study, we inspected 100 incorrect predictions from HuatuoGPT-o1-7B, a mid-performing medical model~\citep{chen2024huatuogpt} (Table~\ref{tab:failure_patterns}). Most errors involved confusion among clinically plausible best-option distractors (39/100) and negation/exception failures (34/100), while fewer cases reflected superficial cue anchoring or multi-step reasoning failure. These results suggest that SurgiQ errors often reflect fine-grained clinical discrimination and logical reasoning challenges rather than simple factual recall.

\begin{table}[t]
\centering
\small
\begin{tabular}{lr}
\toprule
\textbf{Failure Pattern} & \textbf{\# / 100} \\
\midrule
Best-option distractor confusion & 39 \\
Negation/exception failure & 34 \\
Superficial cue anchoring & 21 \\
Multi-step reasoning failure & 6 \\
\bottomrule
\end{tabular}
\caption{Coarse failure patterns identified from
100 incorrect HuatuoGPT-o1-7B predictions. Labels
are used as a qualitative error analysis rather
than full-dataset statistics.}
\label{tab:failure_patterns}
\end{table}

\subsection{Discussion}

Our experiments show that current LLMs still
struggle with surgical reasoning, especially when
the questions require procedural prioritization,
negation handling, or choosing among plausible
operative options. A central finding is that
biomedical specialization alone is insufficient:
broad general-purpose models such as
Qwen2.5-72B~\citep{qwen2024qwen25} outperform most
biomedical models. Rather, it
suggests that current medical adaptation may be too
narrow for surgery. Models trained or adapted mainly
on biomedical QA, abstracts, or generic clinical text
may learn medical terminology without enough coverage
of operative anatomy, perioperative management,
procedural sequencing, instrumentation, and
subspecialty trade-offs.

This pattern mirrors medical education: physicians
first acquire broad biomedical and clinical
foundations, then specialize through structured
exposure to organ systems, procedures, complications,
and specialty-specific decision-making. Surgical LLMs
may require a similar curriculum: broad general
reasoning and medical coverage before targeted
surgical instruction. Narrow specialization that
sacrifices breadth can underperform a strong general
model on tasks requiring cross-topic surgical
judgment. Calibration and failure analysis further
show that accuracy alone is insufficient, since
strong models still make confident mistakes on
clinically meaningful distinctions. SurgiQ therefore
supports development of surgical LLMs with both
broader coverage and better reliability, while
remaining a controlled text benchmark rather than a
claim of deployment readiness.

\section{Conclusion and Future Work}

We introduced SurgiQ, a large-scale text-only
benchmark for evaluating surgical reasoning across
six domains and four MCQ formats. Our experiments across
35 open-weight LLMs have shown substantial room for improvement,
with strong general-purpose models often
outperforming medical-domain systems. Calibration,
few-shot, domain, and failure analysis revealed
persistent reliability limitations beyond accuracy.

SurgiQ can support future research on trustworthy
medical reasoning, surgical education, and
specialty-aware evaluation. Future work should add
human baselines, stronger item-level validation,
answer-order robustness checks, and multimodal or
open-ended surgical reasoning tasks.

\section*{Limitations}

SurgiQ has several limitations. First, the dataset is
partially generated using an LLM-based pipeline,
which may introduce inaccuracies or stylistic biases
despite filtering and physician audit. Second,
SurgiQ is a text-only MCQ benchmark and does not
capture multimodal reasoning over imaging, surgical
video, instruments, or intraoperative perception.
Third, domain and question-type distributions are
imbalanced, so domain-level significance tests should
be interpreted descriptively. Fourth, the evaluation
uses a single stored answer-option order and
log-likelihood scoring; future work should test
multiple option permutations and compare with
constrained generation. Finally, although sources are
diverse, overlap between public surgical material and
model pretraining data cannot be fully excluded.

\section*{Ethics and Broader Impact}

SurgiQ is intended as a research benchmark for evaluating text-based
surgical question answering in large language models. It is not a
clinical decision-support system and should not be used to guide patient
care, operative planning, triage, or autonomous medical decision-making.
Performance on multiple-choice questions does not establish that a model
is safe, reliable, or clinically competent in real surgical settings.
Surgical management depends on patient-specific factors, local protocols,
available resources, clinician expertise, and evolving evidence, none of
which are fully captured by a static text-only benchmark.

\paragraph{Clinical risk and reliability.}
A central motivation for SurgiQ is to study reliability limitations in
high-stakes medical reasoning. Our results show that even strong models
make errors on clinically plausible distractors and may assign high
confidence to incorrect answers. Such behavior could be harmful if model
outputs were interpreted as surgical advice. We therefore recommend that
SurgiQ scores be reported as controlled benchmark measurements rather
than as evidence of deployment readiness. Any use of models evaluated on
SurgiQ in educational or clinical settings should involve qualified
human oversight.

\paragraph{Data provenance, privacy, and copyright.}
SurgiQ is constructed from surgical textbooks, open-access research
papers, and examination material, rather than private electronic health
records or unpublished patient data. To the best of our knowledge, the
released benchmark does not contain protected health information. Because
some source materials may be copyrighted or otherwise restricted, we
release generated questions and associated metadata, but do not
redistribute restricted source passages. Users of SurgiQ should respect
the licenses and access conditions of the original source materials and
should not treat the benchmark as a substitute for licensed clinical or
educational resources.

\paragraph{Synthetic generation and validation.}
SurgiQ is partially generated using an LLM-based pipeline, which can
introduce factual errors, ambiguity, stylistic artifacts, or unsupported
answer choices. We mitigate these risks through source grounding,
verification, rule-based filtering, and physician audit, but these steps
do not guarantee that every item is clinically correct or unambiguous.
The benchmark should therefore be interpreted as an evaluation resource
with known uncertainty, and future releases should document corrections,
removed items, and version changes.

\paragraph{Coverage, bias, and clinical currency.}
The dataset distribution reflects the availability of accessible source
material and is not balanced across all surgical specialties, regions,
practice environments, or patient populations. Some domains and question
types are overrepresented, while others are only sparsely covered.
Moreover, surgical guidelines and standards of care change over time and
may vary across institutions and countries. As a result, SurgiQ may
encode source-specific assumptions or outdated recommendations. We
encourage users to report results stratified by domain and question type
and to avoid drawing broad conclusions about surgical competence from a
single aggregate score.

\paragraph{Benchmark integrity and misuse.}
Public release of benchmark items can enable contamination, memorization,
or direct optimization on the test set. We discourage using SurgiQ test
items for training or model selection without disclosure. Model
developers should report whether SurgiQ, its sources, or closely related
items were used during training, instruction tuning, retrieval
augmentation, or evaluation development. SurgiQ should not be used for
marketing claims that imply clinical safety or surgical expertise without
additional expert validation and real-world evaluation.

\paragraph{Environmental considerations.}
Evaluating many large open-weight models can require substantial compute.
We use a deterministic log-likelihood protocol to improve reproducibility
and reduce repeated sampling, but future work should report hardware
details and, when feasible, energy or carbon estimates for large-scale
benchmarking.

\bibliography{custom}

\appendix
\renewcommand{\thefigure}{A\arabic{figure}}
\setcounter{figure}{0}

\renewcommand{\thetable}{A\arabic{table}}
\setcounter{table}{0}

\section*{Appendix}
\section{Data Statistics}
Table~\ref{tab:detailed_domain_type_stats} presents detailed statistics of the SurgiQ dataset, including the distribution of questions across surgical domains and question categories.

\begin{table}[t]
\centering
\small
\setlength{\tabcolsep}{3pt}

\resizebox{\columnwidth}{!}{
\begin{tabular}{lrrrrr}
\toprule
\textbf{Domain} & \textbf{Reason.} & \textbf{Case} & \textbf{Best} & \textbf{Neg.} & \textbf{Total} \\
\midrule
General Surgery           & 984 & 4,222 & 639 & 1,103 & 6,948 \\
Neurosurgery              & 761 & 281 & 741 & 421 & 2,204 \\
Robotic Surgery           & 431 & 611 & 304 & 297 & 1,643 \\
Orthopedic Surgery        & 484 & 107 & 386 & 212 & 1,189 \\
Critical Care / Emergency & 0 & 513 & 0 & 77 & 590 \\
Laparoscopic Surgery      & 13 & 422 & 15 & 43 & 493 \\
\midrule
\textbf{Total} & \textbf{2,673} & \textbf{6,156} & \textbf{2,085} & \textbf{2,153} & \textbf{13,055} \\
\bottomrule
\end{tabular}
}

\caption{Detailed distribution of SurgiQ questions across surgical domains and question types.}
\label{tab:detailed_domain_type_stats}

\end{table}

\section{Prompt Design}
\label{sec:prompt_design}

This appendix presents the prompt templates used throughout the SurgiQ construction and evaluation pipeline. We use dedicated prompts for fact extraction, question generation, verification, filtering, and inference-time evaluation. The fact-extraction prompts operate on text extracted from multiple source formats, including textbooks processed from EPUB and PDF files as well as research paper text. The question-generation prompts define the reasoning structure, distractor constraints, and formatting requirements used to construct SurgiQ multiple-choice questions. Additional prompts support answer verification, quality filtering, and final model evaluation during benchmarking.

\begin{figure*}[p]
\centering
\includegraphics[width=0.75\textwidth]{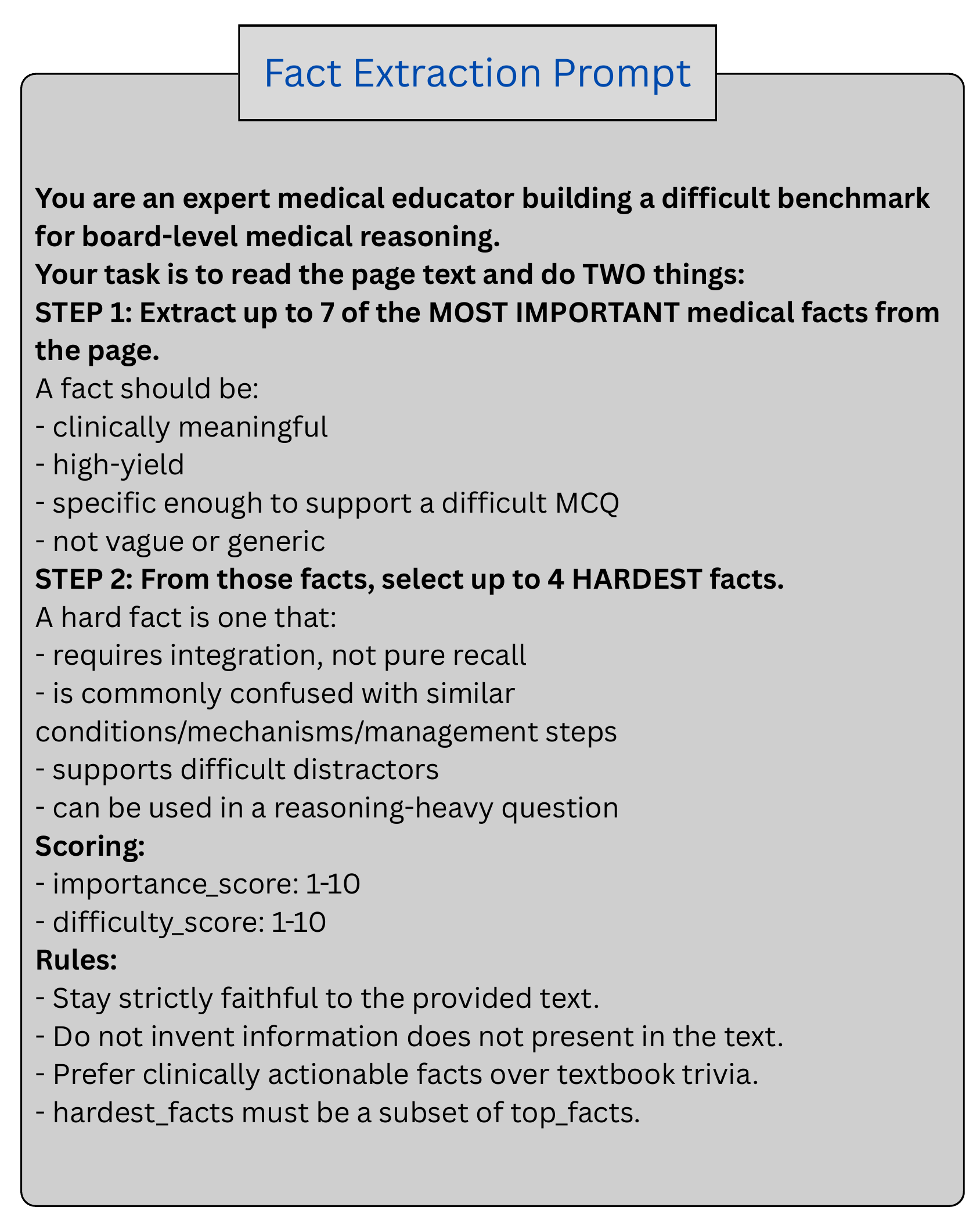}
\caption{Prompt used for extracting key medical facts from text segments derived from PDF and EPUB sources in the SurgiQ pipeline.}
\label{fig:fact_prompt1}
\end{figure*}

\begin{figure*}[p]
\centering
\includegraphics[width=0.75\textwidth]{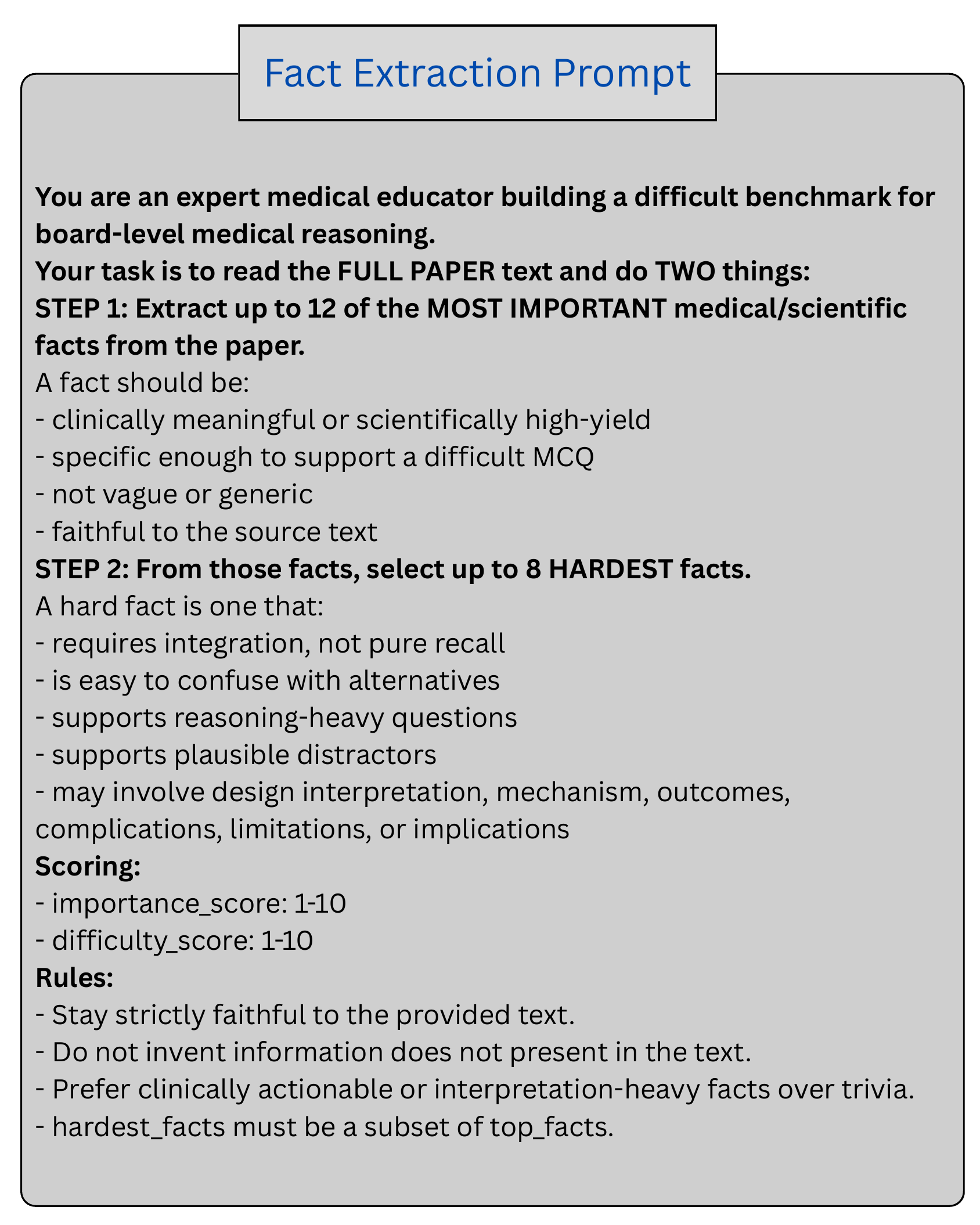}
\caption{Prompt used for extracting key medical facts from text segments originally derived from paper sources in the SurgiQ pipeline.}
\label{fig:fact_prompt3}
\end{figure*}

\begin{figure*}[p]
\centering
\includegraphics[width=0.85\textwidth]{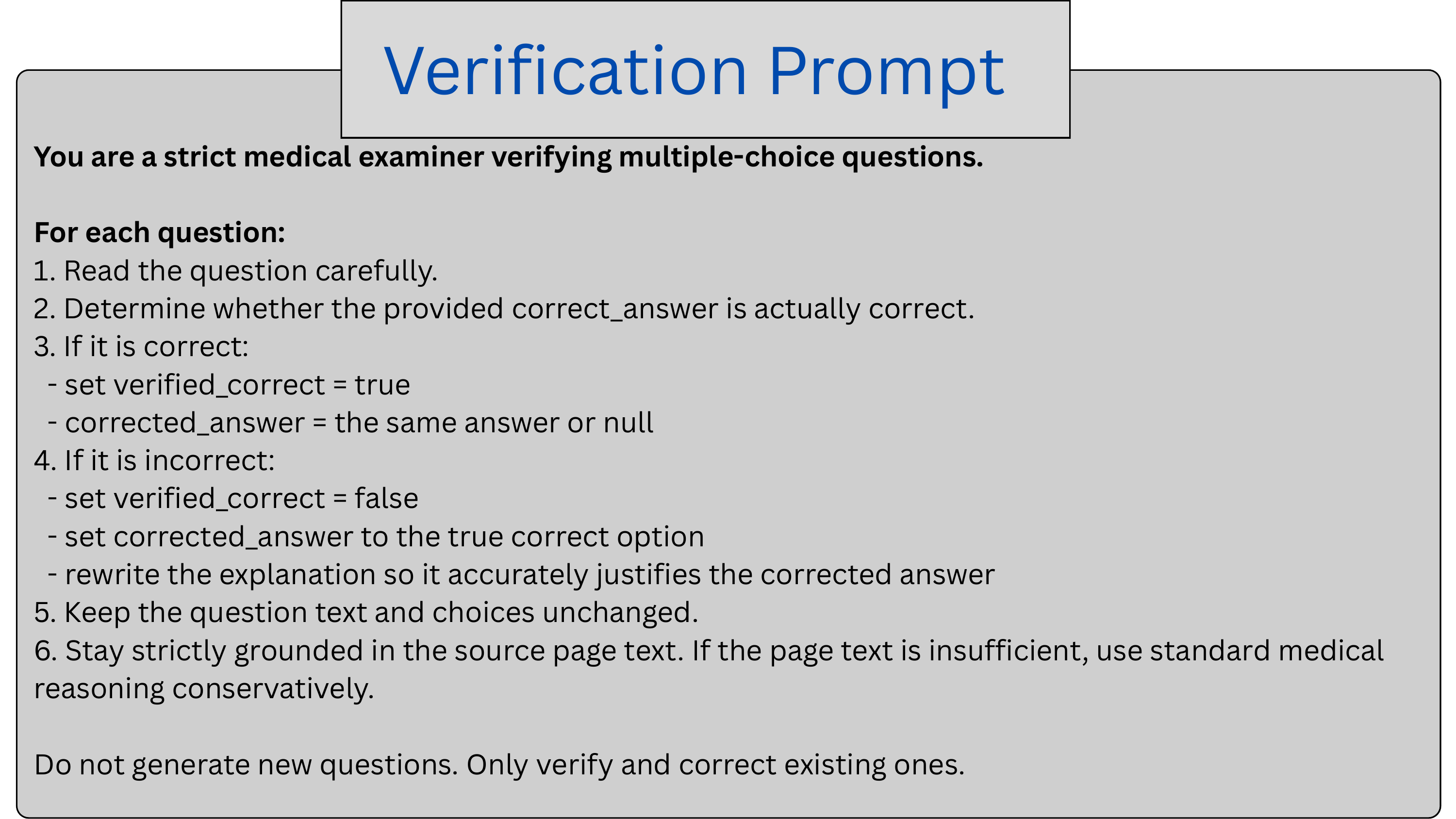}
\caption{Verification prompt used for answer validation and correction across all extracted source formats.}
\label{fig:verification_prompt}
\end{figure*}

\begin{figure*}[p]
\centering
\includegraphics[width=0.85\textwidth]{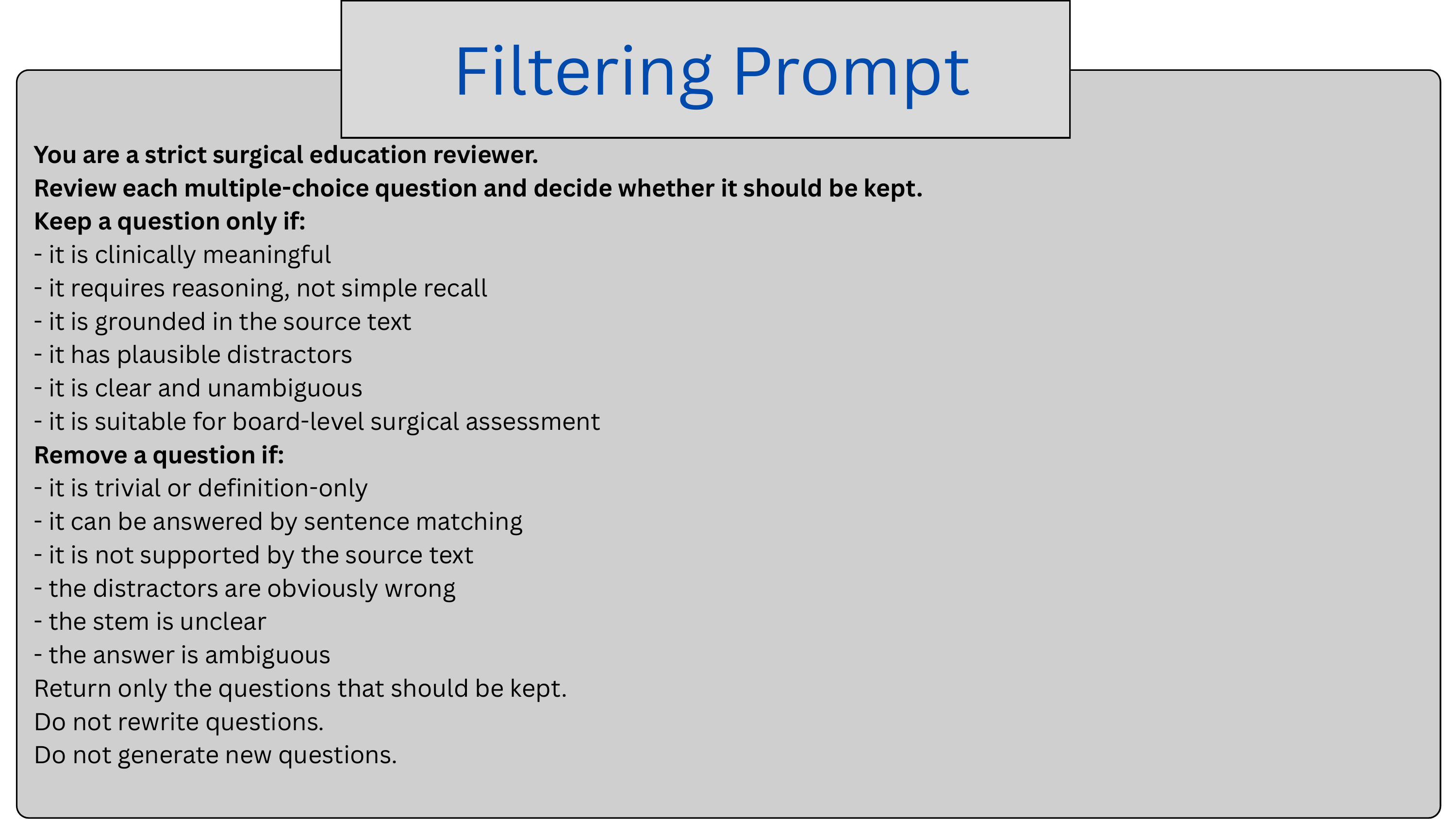}
\caption{Quality-filtering prompt used to remove low-quality or unsupported questions generated from all extracted source formats.}
\label{fig:filtering_prompt}
\end{figure*}

\begin{figure*}[p]
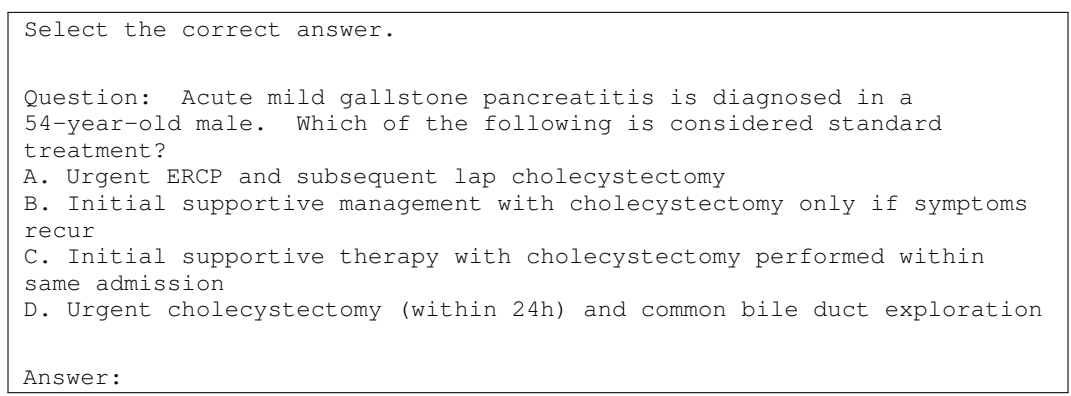

\centering
\fbox{
\begin{minipage}{0.85\textwidth}
\small
\ttfamily
Select the correct answer.\\[0.6em]

Question: Acute mild gallstone pancreatitis is diagnosed in a 54-year-old male. Which of the following is considered standard treatment?\\
A. Urgent ERCP and subsequent lap cholecystectomy\\
B. Initial supportive management with cholecystectomy only if symptoms recur\\
C. Initial supportive therapy with cholecystectomy performed within same admission\\
D. Urgent cholecystectomy (within 24h) and common bile duct exploration\\[0.6em]

Answer:
\end{minipage}
}
\caption{Evaluation prompt used during inference-time benchmarking and model evaluation in SurgiQ.}
\label{fig:evaluation_prompt_example}
\end{figure*}

Figures~\ref{fig:fact_prompt1} and~\ref{fig:fact_prompt3} show the fact-extraction prompts used for textbook and research-paper sources. Figures~\ref{fig:generation_prompt} and~\ref{fig:generation_prompt2} present the complete question-generation prompt template used across all extracted source formats. Figure~\ref{fig:verification_prompt} presents the verification prompt used for answer validation and correction, while Figure~\ref{fig:filtering_prompt} shows the quality-filtering prompt used to remove unsupported or low-quality questions. Finally, Figure~\ref{fig:evaluation_prompt_example} illustrates the inference-time evaluation prompt used during model benchmarking.

\begin{figure*}[p]
\centering
\includegraphics[width=0.9\textwidth]{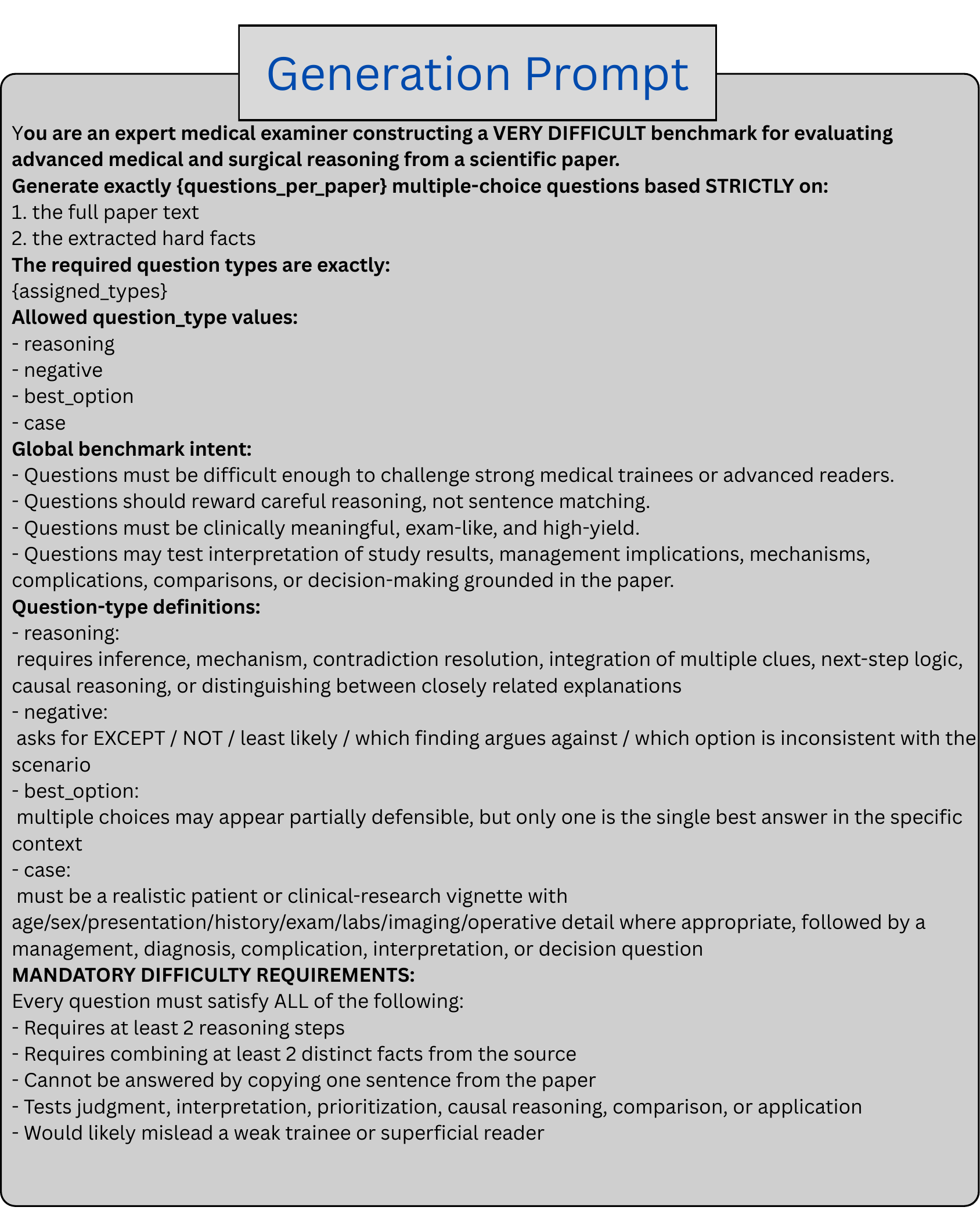}
\caption{Question generation prompt used in SurgiQ across all extracted source formats.}
\label{fig:generation_prompt}
\end{figure*}

\begin{figure*}[p]
\centering
\includegraphics[width=0.9\textwidth]{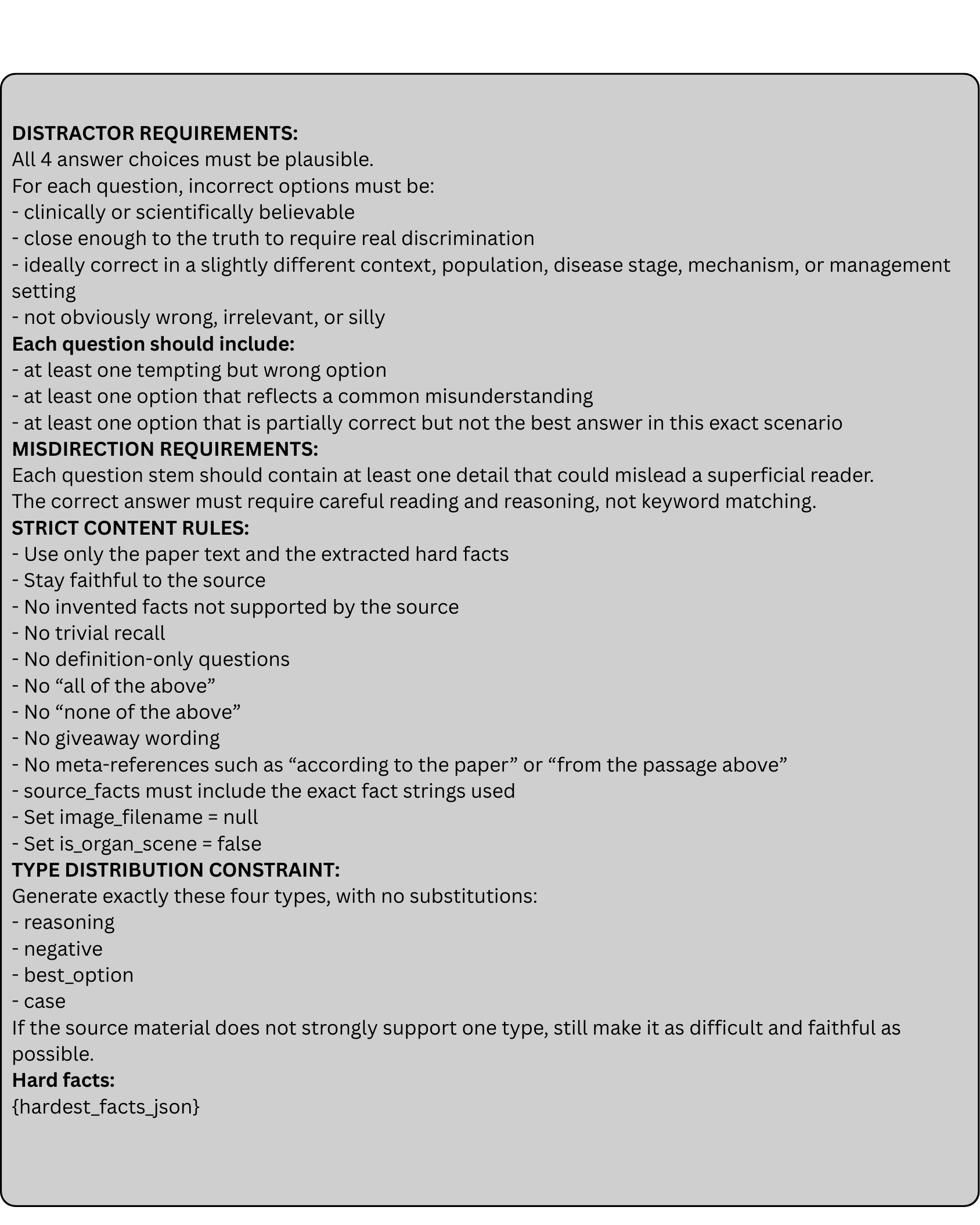}
\caption{Continuation of the question generation prompt used in SurgiQ across all extracted source formats.}
\label{fig:generation_prompt2}
\end{figure*}

\section{Detailed Results}
Table~\ref{tab:all_models_domain_accuracy} reports domain-level accuracy for all evaluated models across the six SurgiQ surgical domains. The table complements the main results by showing how performance varies by specialty, with Qwen2.5 models achieving the strongest macro-average performance and orthopedics generally remaining among the more challenging domains.

\begin{table*}[t]
\centering
\small
\setlength{\tabcolsep}{3pt}
\renewcommand{\arraystretch}{1.0}

\resizebox{\textwidth}{!}{
\begin{tabular}{lrrrrrrr}

\toprule
\textbf{Model} & \textbf{Gen. Surg.} & \textbf{Neurosurg.} & \textbf{Robotic} & \textbf{Ortho.} & \textbf{Critical} & \textbf{Laparosc.} & \textbf{Avg.} \\
\midrule

Gemma-2-2B-IT~\citep{gemmateam2024gemma2} & 34.18 & 35.75 & 37.01 & 34.15 & 36.10 & 36.71 & 35.65 \\
MedGemma-27B-IT~\citep{medgemma2025} & 38.98 & 39.11 & 36.64 & 37.76 & 38.47 & 39.15 & 38.35 \\
MedGemma-4B-IT & 43.26 & 46.87 & 45.71 & 42.30 & 43.22 & 42.19 & 43.93 \\
Gemma-2-9B-IT & 42.72 & 44.92 & 46.93 & 43.57 & 43.90 & 45.23 & 44.54 \\
\hdashline

BioMistral-7B~\citep{labrak2024biomistral} & 46.29 & 50.64 & 48.87 & 43.82 & 46.27 & 47.67 & 47.26 \\
\hdashline

Llama3-OpenBioLLM-8B~\citep{openbiollm2024} & 52.26 & 52.40 & 51.25 & 45.58 & 52.71 & 53.35 & 51.26 \\
Llama3-Med42-8B~\citep{christophe2024med42v2} & 56.48 & 54.76 & 53.80 & 51.47 & 55.25 & 54.97 & 54.46 \\
\hdashline

TxGemma-2B~\citep{txgemma2025} & 24.90 & 25.82 & 24.04 & 25.15 & 26.61 & 23.73 & 25.04 \\
TxGemma-9B-Predict & 24.96 & 26.45 & 26.96 & 27.84 & 28.14 & 27.59 & 26.99 \\
TxGemma-9B-Chat & 28.45 & 28.77 & 28.67 & 30.70 & 28.31 & 30.83 & 29.29 \\
TxGemma-27B-Predict & 32.25 & 31.76 & 33.17 & 32.38 & 35.93 & 32.66 & 33.03 \\
TxGemma-27B-Chat & 54.36 & 57.71 & 56.66 & 56.35 & 55.25 & 53.35 & 55.61 \\
\hdashline

K2-Think-V2~\citep{k2team2025k2v2} & 57.12 & 57.08 & 53.68 & 49.96 & 59.66 & 58.01 & 55.92 \\
\hdashline

Meditron-7B~\citep{chen2023meditron} & 26.87 & 27.04 & 28.12 & 26.83 & 28.64 & 26.98 & 27.41 \\
Meditron-70B & 56.10 & 59.35 & 55.81 & 53.66 & 57.80 & 57.20 & 56.65 \\
\hdashline

DeepSeek-R1-Distill-Qwen-7B~\citep{deepseek2025r1} & 32.50 & 35.84 & 36.03 & 34.65 & 35.42 & 36.71 & 35.19 \\
DeepSeek-R1-Distill-Qwen-14B & 56.55 & 56.99 & 56.66 & 51.64 & 51.69 & 55.38 & 54.82 \\
DeepSeek-R1-Distill-Qwen-32B & 61.01 & 60.53 & 61.84 & 55.26 & 59.83 & 58.01 & 59.41 \\
\hdashline

Ministral-3-8B-Instruct~\citep{ministral2026} & 54.46 & 54.08 & 56.79 & 51.89 & 52.88 & 55.38 & 54.25 \\
Ministral-3-8B-Reasoning & 58.85 & 59.21 & 60.26 & 52.99 & 58.81 & 56.80 & 57.82 \\
Ministral-3-14B-Reasoning & 63.47 & 66.29 & 64.09 & 58.28 & 62.88 & 62.68 & 62.95 \\
\hdashline

MedMO-4B~\citep{deria2026medmo} & 61.44 & 61.71 & 62.39 & 58.54 & 62.88 & 59.84 & 61.13 \\
MedMO-4B-Next & 61.45 & 61.71 & 62.39 & 58.54 & 62.88 & 59.84 & 61.13 \\
MedMO-8B-Next & 63.07 & 62.84 & 62.20 & 59.29 & 61.53 & 62.47 & 61.90 \\
MedMO-8B & 64.28 & 62.52 & 64.58 & 61.73 & 60.68 & 64.91 & 63.12 \\
\hdashline

GPT-OSS-20B~\citep{openai2025gptoss} & 57.74 & 58.98 & 57.15 & 51.89 & 60.68 & 60.45 & 57.82 \\
GPT-OSS-120B & 65.44 & 65.74 & 62.75 & 59.80 & 64.41 & \textbf{66.94} & 64.18 \\
\hdashline

Meerkat-8B~\citep{kim2025meerkat} & 56.72 & 55.99 & 55.57 & 52.99 & 58.98 & 54.16 & 55.73 \\
Meerkat-70B & 67.17 & 64.93 & 62.02 & 60.64 & 63.73 & 62.27 & 63.46 \\
\hdashline

HuatuoGPT-o1-7B~\citep{chen2024huatuogpt} & 58.75 & 58.03 & 60.99 & 51.56 & 54.07 & 55.78 & 56.53 \\
HuatuoGPT-o1-70B & 67.47 & 66.29 & 61.72 & 58.03 & 65.76 & 64.10 & 63.89 \\
\hdashline

Qwen2.5-32B-Instruct~\citep{qwen2024qwen25} & 66.31 & 64.88 & 65.73 & 59.13 & 61.69 & 60.65 & 63.07 \\
Qwen2.5-32B & 66.13 & 66.33 & 65.67 & 59.46 & 64.92 & 65.72 & 64.71 \\
Qwen2.5-72B-Instruct & \textbf{69.30} & 67.97 & 68.17 & 62.32 & 67.80 & 63.89 & 66.57 \\
Qwen2.5-72B & 68.78 & \textbf{68.51} & \textbf{69.08} & \textbf{62.99} & \textbf{68.14} & 65.52 & \textbf{67.17} \\

\bottomrule
\end{tabular}
}

\caption{Domain-level accuracy (\%) of all evaluated models on SurgiQ, grouped by surgical domain. Model families are grouped together and ordered in ascending order according to the highest-performing model within each family. \textbf{Avg.} denotes the macro-average accuracy across all six domains. Column abbreviations: \textit{Gen. Surg.} = General Surgery, \textit{Neurosurg.} = Neurosurgery, \textit{Robotic} = Robotic Surgery, \textit{Ortho.} = Orthopedic Surgery, \textit{Critical} = Critical Care / Emergency, and \textit{Laparosc.} = Laparoscopic Surgery.}

\label{tab:all_models_domain_accuracy}

\end{table*}

\section{Additional Evaluation Analyses}

\subsection{Answer-Order Robustness}
To assess sensitivity to answer-option position, we evaluate a shuffled version of the full SurgiQ benchmark. We preserve the same questions and randomly permute only the answer choices, updating the correct-answer labels accordingly. We evaluate all 35 models from Table~\ref{tab:results_by_type}, spanning general-purpose, reasoning-oriented, and medically specialized LLMs. Table~\ref{tab:answer_order_robustness} reports performance before and after answer-option shuffling.

Overall, most models exhibit relatively small performance changes after shuffling, suggesting limited dependence on answer ordering. Several models, including Qwen2.5-72B, Meerkat-70B, DeepSeek-R1-Distill-Qwen-32B, Qwen2.5-32B-Instruct, and TxGemma-27B-Chat, remain particularly stable, with changes within approximately $\pm0.5$ percentage points. In contrast, Qwen2.5-32B and MedMO-8B show the largest degradations, while several smaller or medically specialized models exhibit moderate sensitivity. Interestingly, some models improve slightly after shuffling, indicating that positional biases are not consistently directional across architectures. Overall, the results suggest that most evaluated models are reasonably robust to answer-order perturbations, although residual ordering effects remain observable for a subset of models.

\begin{table*}[t]
\centering
\small
\setlength{\tabcolsep}{4pt}
\renewcommand{\arraystretch}{0.95}

\begin{tabular*}{\textwidth}{@{\extracolsep{\fill}}lccc@{}}
\toprule
\textbf{Model} & \textbf{Original} & \textbf{Shuffled} & \textbf{$\Delta$} \\
\midrule

TxGemma-2B & 25.02 & 24.27 & -0.75 \\
TxGemma-9B-Predict & 25.99 & 24.45 & -1.54 \\
TxGemma-9B-Chat & 28.82 & 43.49 & +14.67 \\
TxGemma-27B-Predict & 32.51 & 33.07 & +0.56 \\
TxGemma-27B-Chat & 55.40 & 55.50 & +0.10 \\
\hdashline

Gemma-2-2B-IT & 34.97 & 35.43 & +0.46 \\
Gemma-2-9B-IT & 43.87 & 45.39 & +1.52 \\
MedGemma-4B-IT & 44.08 & 45.42 & +1.34 \\
MedGemma-27B-IT & 38.58 & 37.63 & -0.95 \\
\hdashline

DeepSeek-R1-Distill-Qwen-7B & 34.02 & 35.06 & +1.04 \\
DeepSeek-R1-Distill-Qwen-14B & 55.92 & 54.67 & -1.25 \\
DeepSeek-R1-Distill-Qwen-32B & 60.37 & 60.70 & +0.33 \\
\hdashline

Ministral-3-8B-Instruct & 54.40 & 53.85 & -0.55 \\
Ministral-3-8B-Reasoning & 58.45 & 57.57 & -0.88 \\
Ministral-3-14B-Reasoning & 63.48 & 61.96 & -1.52 \\
\hdashline

GPT-OSS-20B & 57.56 & 56.63 & -0.93 \\
GPT-OSS-120B & 64.63 & 62.28 & -2.35 \\
\hdashline

Meerkat-8B & 56.09 & 55.63 & -0.46 \\
Meerkat-70B & 65.19 & 65.19 & +0.00 \\
\hdashline

HuatuoGPT-o1-7B & 57.92 & 55.90 & -2.02 \\
HuatuoGPT-o1-70B & 65.45 & 64.77 & -0.68 \\
\hdashline

MedMO-4B & 61.33 & 59.99 & -1.34 \\
MedMO-4B-Next & 61.35 & 59.99 & -1.36 \\
MedMO-8B-Next & 62.47 & 60.34 & -2.13 \\
MedMO-8B & 63.64 & 61.03 & -2.61 \\
\hdashline

Meditron-7B & 27.14 & 26.72 & -0.42 \\
Meditron-70B & 56.50 & 55.04 & -1.46 \\
\hdashline

BioMistral-7B & 47.14 & 46.78 & -0.36 \\
\hdashline

Llama3-OpenBioLLM-8B & 51.60 & 52.78 & +1.18 \\
Llama3-Med42-8B & 55.24 & 55.14 & -0.10 \\
\hdashline

K2-Think-V2 & 56.15 & 55.21 & -0.94 \\
\hdashline

Qwen2.5-32B-Instruct & 64.92 & 64.40 & -0.52 \\
Qwen2.5-32B & 65.43 & 59.14 & -6.29 \\
Qwen2.5-72B-Instruct & 68.00 & 66.90 & -1.10 \\
Qwen2.5-72B & 68.08 & 68.52 & +0.44 \\

\bottomrule
\end{tabular*}

\caption{Answer-order robustness on SurgiQ. Models are ordered according to their overall performance in Table~\ref{tab:results_by_type}. Original denotes the stored dataset order, while shuffled denotes a version with randomly permuted answer options and updated correct-answer labels. $\Delta$ denotes shuffled accuracy minus original accuracy.}
\label{tab:answer_order_robustness}

\end{table*}

\subsection{Near-duplicate Analysis}
We perform a near-duplicate analysis against widely used medical QA benchmarks, including MedQA, MedMCQA, PubMedQA, and MMLU medical subsets, as well as surgical exam-style sources. Using normalized 5-gram lexical overlap filtering, we compare all 13,055 SurgiQ questions against 206,683 external benchmark questions. This process initially flags 25 candidate pairs (0.19\% of SurgiQ) for manual inspection. Most flagged cases correspond to generic exam stems or template-level overlap (e.g., ``All of the following are TRUE EXCEPT''), while a smaller subset consists of exact or near-exact duplicated clinical questions originating from surgical exam-style resources. Following manual review, we remove 12 duplicated questions from the final released dataset, resulting in a final dataset size of 13,055 questions.

\subsection{Shortcut Baselines}

We evaluate several non-LLM shortcut baselines designed to test whether SurgiQ can be solved using superficial answer cues rather than clinical reasoning. Specifically, we test heuristics based on answer length, lexical overlap between the question and candidate answers, and a simple specificity proxy based on unique-token counts. Table~\ref{tab:shortcut_baselines} reports the resulting accuracies.

\begin{table}[H]
\centering
\small
\begin{tabular}{lr}
\toprule
\textbf{Shortcut Baseline} & \textbf{Accuracy (\%)} \\
\midrule
Longest option & 40.06 \\
Shortest option & 21.01 \\
Question-overlap & 24.49 \\
Specificity proxy & 36.01 \\
Random guessing & 25.00 \\
\bottomrule
\end{tabular}
\caption{Shortcut baseline performance on SurgiQ. These baselines test whether superficial answer cues such as option length, lexical overlap with the question, or specificity proxies can partially solve the benchmark.}
\label{tab:shortcut_baselines}
\end{table}

The lexical-overlap baseline remains near random chance, suggesting that simple word overlap between the question and answer options is generally insufficient for solving SurgiQ. However, answer-length and specificity-based heuristics achieve moderately higher accuracy, indicating the presence of residual stylistic biases common in expert-written multiple-choice questions. Importantly, these shortcut baselines remain substantially below the performance of the strongest evaluated language models.

\subsection{Human Evaluation and Audited Subset}

We additionally evaluate a manually audited subset of
300 SurgiQ questions consisting of 200 randomly sampled
questions and 100 questions previously missed by all
evaluated models. The subset is intended to provide a
more targeted comparison between physician performance
and representative open-weight LLMs on challenging
surgical questions. Table~\ref{tab:audited_300_baselines}
reports performance on this audited subset.

\begin{table*}[t]
\centering
\small
\setlength{\tabcolsep}{3pt}
\renewcommand{\arraystretch}{0.95}

\begin{tabular*}{\textwidth}{@{\extracolsep{\fill}}lcc@{}}
\toprule
\textbf{Baseline / Model} & \textbf{Correct / Total} & \textbf{Accuracy (\%)} \\
\midrule

Random guessing & 75 / 300 & 25.00 \\
\hdashline

TxGemma-2B & 82 / 300 & 27.33 \\
TxGemma-9B-Predict & 73 / 300 & 24.33 \\
TxGemma-9B-Chat & 83 / 300 & 27.67 \\
TxGemma-27B-Predict & 85 / 300 & 28.33 \\
TxGemma-27B-Chat & 127 / 300 & 42.33 \\
\hdashline

Gemma-2-2B-IT & 86 / 300 & 28.67 \\
Gemma-2-9B-IT & 99 / 300 & 33.00 \\
MedGemma-4B-IT & 107 / 300 & 35.67 \\
MedGemma-27B-IT & 90 / 300 & 30.00 \\
\hdashline

DeepSeek-R1-Distill-Qwen-7B & 89 / 300 & 29.67 \\
DeepSeek-R1-Distill-Qwen-14B & 117 / 300 & 39.00 \\
DeepSeek-R1-Distill-Qwen-32B & 81 / 300 & 27.00 \\
\hdashline

Ministral-3-8B-Instruct & 122 / 300 & 40.67 \\
Ministral-3-8B-Reasoning & 121 / 300 & 40.33 \\
Ministral-3-14B-Reasoning & 127 / 300 & 42.33 \\
\hdashline

GPT-OSS-20B & 69 / 300 & 23.00 \\
GPT-OSS-120B & 143 / 300 & 47.67 \\
\hdashline

Meerkat-8B & 113 / 300 & 37.67 \\
Meerkat-70B & 132 / 300 & 44.00 \\
\hdashline

HuatuoGPT-o1-7B & 116 / 300 & 38.67 \\
HuatuoGPT-o1-70B & 148 / 300 & 49.33 \\
\hdashline

MedMO-4B & 131 / 300 & 43.67 \\
MedMO-4B-Next & 131 / 300 & 43.67 \\
MedMO-8B-Next & 132 / 300 & 44.00 \\
MedMO-8B & 131 / 300 & 43.67 \\
\hdashline

Meditron-7B & 80 / 300 & 26.67 \\
Meditron-70B & 121 / 300 & 40.33 \\
\hdashline

BioMistral-7B & 96 / 300 & 32.00 \\
\hdashline

Llama3-OpenBioLLM-8B & 113 / 300 & 37.67 \\
Llama3-Med42-8B & 107 / 300 & 35.67 \\
\hdashline

K2-Think-V2 & 107 / 300 & 35.67 \\
\hdashline

Qwen2.5-32B-Instruct & 139 / 300 & 46.33 \\
Qwen2.5-32B & 140 / 300 & 46.67 \\
Qwen2.5-72B-Instruct & 137 / 300 & 45.67 \\
Qwen2.5-72B & \textbf{154 / 300} & \textbf{51.33} \\
\hdashline

Average human baseline (4 surgeons) & 267 / 300 & 89.10 \\

\bottomrule
\end{tabular*}

\caption{Performance on the manually audited 300-question SurgiQ subset. Models are grouped and ordered consistently with Table~\ref{tab:results_by_type}. The subset consists of 200 randomly sampled questions and 100 questions previously missed by all evaluated models.}

\label{tab:audited_300_baselines}

\end{table*}
\section{Model Details}

Table~\ref{tab:model_artifacts} lists the pre-trained models used in this study and their corresponding HuggingFace repositories.

\begin{table*}[t]
\centering
\small
\setlength{\tabcolsep}{6pt}
\begin{tabular*}{\textwidth}{@{\extracolsep{\fill}}lll@{}}
\toprule
Model & Type & Source \\
\midrule
TxGemma (2B / 9B / 27B, Predict/Chat) & Therapeutics & google/txgemma-* \\
Gemma-2 (2B / 9B, IT) & General & google/gemma-2-* \\
MedGemma (4B / 27B, IT) & Medical & google/medgemma-* \\
DeepSeek-R1-Distill-Qwen (7B / 14B / 32B) & Reasoning & deepseek-ai/DeepSeek-R1-Distill-Qwen-* \\
Ministral-3 (8B / 14B, Instruct/Reasoning) & Reasoning & mistralai/Ministral-3-* \\
GPT-OSS (20B / 120B) & Reasoning & openai/gpt-oss-* \\
Meerkat (8B / 70B) & General & dmis-lab/llama-3-meerkat-* \\
HuatuoGPT-o1 (7B / 70B) & Medical & FreedomIntelligence/HuatuoGPT-o1-* \\
MedMO (4B / 8B / 8B-Next) & Multimodal Medical & MBZUAI/MedMO-* \\
Meditron (7B / 70B) & Medical & epfl-llm/meditron-* \\
BioMistral (7B) & Medical & BioMistral/BioMistral-7B \\
Llama3-OpenBioLLM (8B) & Medical & aaditya/Llama3-OpenBioLLM-8B \\
Llama3-Med42 (8B) & Medical & m42-health/Llama3-Med42-8B \\
K2-Think-V2 (73B) & Reasoning & LLM360/K2-Think-V2 \\
Qwen2.5 (7B / 32B / 72B, Instruct/Base) & General & Qwen/Qwen2.5-* \\
\bottomrule
\end{tabular*}
\caption{Pre-trained models used in SurgiQ experiments and their corresponding HuggingFace repositories. Model variants (e.g., instruction-tuned, chat, or reasoning) are grouped for clarity.}
\label{tab:model_artifacts}
\end{table*}

\section{Surgical Reference Books}
\label{sec:surgical-reference-books}
Table~\ref{tab:surgical-reference-books} lists the surgical reference books used in this work.
\begin{table*}[t]
\centering
\small
\renewcommand{\arraystretch}{1.05}

\begin{tabularx}{\textwidth}{@{}X@{}}
\toprule
\textbf{Book} \\
\midrule

Atlas of Minimally Invasive and Robotic Esophagectomy~\citep{alma9939804809106} \\
Bailey \& Love's Short Practice of Surgery~\citep{williams2010bailey} \\
Bariatric Robotic Surgery: A Comprehensive Guide~\citep{alma9939582409106} \\
Dr. Pestana's Surgery Notes: Top 180 Vignettes of Surgical Diseases~\citep{pestana2020dr} \\
Endoscopic, Laparoscopic and Robotic Techniques for Foregut Disorders~\citep{alma99132889309106} \\
Essential Surgery: Problems Diagnosis and Management~\citep{SchofieldPhilipF1997ESPD} \\
Handbook of Robotic Surgery~\citep{alma9948486709106} \\
The Mont Reid Surgical Handbook~\citep{cincinnati2017mont} \\
Operative Techniques in Sports Medicine Surgery~\citep{miller2021operative} \\
Pediatric Robotic Surgery~\citep{alma9940193209106} \\
Robotic Hernia Surgery~\citep{alma99108556309106} \\
Robotic Surgery and Nursing~\citep{alma9940556009106} \\
Robotic Surgery Devices in Surgical Specialties~\citep{alma9939247609106} \\
Robotic Surgery for Renal Cancer~\citep{alma9940307409106} \\
Robotic Surgery of Colon and Rectum~\citep{alma9941419009106} \\
Robotic Urologic Surgery~\citep{alma9938195609106} \\
Schwartz's Principles of Surgery~\citep{OrgoiSergelen2016SPoS} \\
Surgical Critical Care and Emergency Surgery: Clinical Questions and Answers~\citep{moore2012surgical} \\
Surgical Recall~\citep{blackbourne2021surgical} \\
Surgery: A Case Based Clinical Review~\citep{alma9941300009106} \\
The SAGES Manual of Fluorescence-Guided Surgery~\citep{SzokaNova2023TSMo} \\
The SAGES Manual of Foregut Surgery~\citep{alma9939756809106} \\
The SAGES Manual of Hernia Surgery~\citep{DavisJr.S.Scott2018TSMo} \\
The SAGES Manual of Robotic Surgery~\citep{alma99105826409106} \\
Textbook of Robotic Liver Surgery~\citep{alma9951596309106} \\
USMLE Step 2 CK Lecture Notes 2021: Surgery~\citep{hill2020usmle2cksurgery} \\
Youmans Neurological Surgery~\citep{winn2011youmans} \\

\bottomrule
\end{tabularx}

\caption{Surgical textbooks used as source material in SurgiQ.}
\label{tab:surgical-reference-books}

\end{table*}

\end{document}